\documentclass[sigconf]{acmart}
\usepackage{booktabs} 
\usepackage{subfigure}
\usepackage{bm}
\usepackage{soul}
\usepackage{enumitem}
\usepackage{algorithm}
\usepackage{algorithmic}
\usepackage{makecell}
\usepackage{diagbox}
\usepackage{multirow}
\usepackage{float}
\usepackage{mathrsfs}
\usepackage{colortbl}
\setul{0.5ex}{0.3ex}
\setulcolor{blue}

\newcommand{\myref}[1]{Eq.(\ref{#1})}
\newcommand{\mytbref}[1]{Table \ref{#1}}
\newcommand{\newcite}[1]{\citeauthor{#1} [\citenum{#1}]}

\begin{document}

\settopmatter{printacmref=true}
\fancyhead{}

\title{EnsembleGAN: Adversarial Learning for Retrieval-Generation \\Ensemble Model on Short-Text Conversation}



\author{Jiayi Zhang}
\authornote{Equal contribution.}
\affiliation{%
  \institution{Turing Robot}
}
\email{zhangjiayi@uzoo.cn}

\author{Chongyang Tao}
\authornotemark[1]
\affiliation{%
  \institution{ICST, Peking University}
}
\email{chongyangtao@pku.edu.cn}

\author{Zhenjing Xu}
\affiliation{%
  \institution{Turing Robot}
}
\email{xuzhenjing@uzoo.cn}

\author{Qiaojing Xie}
\affiliation{%
  \institution{Turing Robot}
}
\email{xieqiaojing@uzoo.cn}

\author{Wei Chen}
\affiliation{%
  \institution{Turing Robot}
}
\email{weichen@uzoo.cn}

\author{Rui Yan}
\authornote{Corresponding author: Rui Yan (ruiyan@pku.edu.cn).}
\affiliation{%
  \institution{ICST, Peking University}
}
\email{ruiyan@pku.edu.cn}

\begin{abstract}
Generating qualitative responses has always been a challenge for human-computer dialogue systems. Existing dialogue systems generally derive from either retrieval-based or generative-based approaches, both of which have their own pros and cons. Despite the natural idea of an ensemble model of the two, existing ensemble methods only focused on leveraging one approach to enhance another, we argue however that they can be further \emph{mutually} enhanced with a proper training strategy. In this paper, we propose ensembleGAN, an adversarial learning framework for enhancing a retrieval-generation ensemble model in open-domain conversation scenario. It consists of a language-model-like generator, a ranker generator, and one ranker discriminator. Aiming at generating responses that approximate the ground-truth and receive high ranking scores from the discriminator, the two generators learn to generate improved highly relevant responses and competitive unobserved candidates respectively, while the discriminative ranker is trained to identify true responses from adversarial ones, thus featuring the merits of both generator counterparts. The experimental results on a large short-text conversation data demonstrate the effectiveness of the ensembleGAN by the amelioration on both human and automatic evaluation metrics.
\end{abstract}

%
%

\ccsdesc[300]{Information systems}
\ccsdesc[300]{Information systems~Web applications}
\ccsdesc[300]{Information systems~Retrieval models and ranking}

\copyrightyear{2019}
\acmYear{2019}
\setcopyright{acmcopyright}
\acmConference[SIGIR '19]{Proceedings of the 42nd International ACM SIGIR Conference on Research and Development in Information Retrieval}{July 21--25, 2019}{Paris, France}
\acmBooktitle{Proceedings of the 42nd International ACM SIGIR Conference on
Research and Development in Information Retrieval (SIGIR '19), July 21--25, 2019,
Paris, France}
\acmPrice{15.00}
\acmDOI{10.1145/3331184.3331193}
\acmISBN{978-1-4503-6172-9/19/07}

\keywords{Generative adversarial network, short-text conversation, ensemble method, retrieval-based conversation, generation-based conversation}


\maketitle

\section{Introduction}
Natural language human-computer conversation has long been an attractive but challenging task in artificial intelligence (AI), for it requires both language understanding and reasoning ~\cite{NRM}. While early works mainly focused on domain-specific scenarios such as ticket booking, the open-domain chatbot-human conversations has gained popularity recently, not only for their commercial values (e.g., Xiaoice\footnote[1]{http://www.msxiaoice.com/} from Microsoft), but for the rapid growth of online social media as well, along with tremendous data available for data-driven deep learning methods to be proved worthwhile. Current conversation systems could be generally divided into two different categories, namely the retrieval-based and the generative-based approach.

Given an user input utterance (also called a query), a retrieval-based system usually retrieves a number of response candidates from a pre-constructed index, and then selects the best matching one as a response to a human input using semantic matching~\cite{DBLP:conf/emnlp/WangLLC13,YanSW16,wu2017sequential}. The retrieved responses usually have various expressions with rich information and language fluency. However, limited by the capacity of the pre-constructed repository, the selected response might seem less customized for unobserved novel queries .

Meanwhile the generative conversation system works differently, for it generates responses token by token according to conditional probabilistic language models (LM) such as seq2seq with attention~\cite{DBLP:journals/corr/BahdanauCB14}, which generates appropriate and tailored responses to most queries, but often suffers from the lack of language fluency and the problem of universal responses (e.g., ``I don't know" and ``Me too") due to statistical model incapabilities~\cite{DBLP:journals/sigkdd/ChenLYT17}. Various ameliorations have been proposed to enrich the generation, either by better exploring internal features such as mutual-information-based objective function~\cite{DBLP:conf/naacl/LiGBGD16}, dynamic vocabularies~\cite{DBLP:conf/aaai/WuWYXL18} and diverse beam search~\cite{DBLP:journals/corr/VijayakumarCSSL16}, or by incorporating external knowledge, such as topic information~\cite{DBLP:journals/corr/XingWWLHZM16}, cue words~\cite{DBLP:conf/coling/MouSYL0J16,yao2017towards}, dialog acts~\cite{zhao2017learning}, and common sense knowledge~\cite{young2018augmenting}. 

On the other hand, studies seeking for an ensemble of both retrieval and generative approaches show great improvement to dialogue generation performance. \newcite{DBLP:conf/ijcai/SongLNZZY18} proposed Multi-Seq2Seq model that focuses on leveraging responses generated by the retrieval-based dialog systems to enhance generation-based dialog systems, thus synthesizing a more informative response. Similarly, \newcite{weston2018retrieve} designed a retrieval-and-refine model which treats the retrieval as additional context for sequence generator to avoid universal issues such as producing short sentences with frequent words. \newcite{DBLP:journals/corr/abs-1806-07042} introduced a prototype-then-edit paradigm for their conversation system by building a retrieval-based prototype editing with a seq2seq model that increases the diversity and informativeness
of the generation results. 

Despite the performance gain of an ensemble compared with either retrieval or generative model, previous works only focused on ameliorating one approach based on the other, still leaving great potentials for making further progress by allowing both methods to be mutually enhanced. Inspired by adversarial learning~\cite{DBLP:conf/nips/GoodfellowPMXWOCB14}, we propose a generative adversarial framework for improving an ensemble on short-text conversation, which is called EnsembleGAN throughout the paper. Particularly, EnsembleGAN consists of two generators and a discriminator. The LM-like generator ($G_1$) is responsible for synthesizing tailored responses via a sequence-to-sequence framework, while the ranking-based generator ($G_2$) aims at selecting highly competitive negative responses from a pre-retrieval module and $G_1$, and finally the ranking-based discriminator ($D$) endeavors to
distinguish the ground-truth and adversarial candidates provided by pre-retrieval module and two generators ($G_1$ and $G_2$). 


The motivation behind is that through adversarial learning, with $G_1$ generating improved highly relevant responses, and $G_2$ providing enriched and fluent unobserved as well as synthetic candidates, the discriminative ranker could be further trained to identify responses that are highly correlated, informative and fluent, thus absorbing the merits of both its generative counterparts. The proposed EnsembleGAN framework is intuitively suited for improving a combination of any neural-based generative and retrieval approaches towards better global optimal results. The main contribution of this paper is three-folded and it's summarized as follows:
\begin{itemize}
\item We introduce a novel end-to-end generative adversarial framework that aims to mutually enhance both generative and retrieval module, leading to a better amelioration of a dialogue ensemble model.
\item We make extensive studies on ensembles of various generators and discriminators, providing insights of global and local optimization from the ensemble perspective through both quantitative and qualitative analysis.
\item We demonstrate the effectiveness of the proposed EnsembleGAN by performing experiments on a large mixed STC dataset, the gain on various metrics confirms that the ensemble model as well as each of its modules could all be enhanced by our method.
\end{itemize}

\section{Related Work}
Open-domain dialogue systems have been attracting increasing attention in recent years. Researchers have made various progress on building both generative-based~\cite{NRM,DBLP:conf/coling/MouSYL0J16,tao2018get} and retrieval-based conversation system~\cite{YanSW16,yan2017joint,wu2017sequential,yan2018coupled,tao2019multi}. Besides, with the success of generative adversarial networks (GANs)~\cite{DBLP:conf/nips/GoodfellowPMXWOCB14} on computer vision such as image translation~\cite{DBLP:journals/corr/ZhuPIE17} and image captioning~\cite{DBLP:conf/iccv/DaiFUL17}, studies of GAN applications also start to emerge in the domain of natural language processing (NLP), such as dialogue generation~\cite{DBLP:conf/emnlp/LiMSJRJ17,DBLP:journals/corr/abs-1802-01345}, machine translation~\cite{DBLP:conf/naacl/YangCWX18} and text summarization~\cite{DBLP:conf/aaai/LiuLYQZL18}, all demonstrating the effectiveness of GAN mechanism in the domain of NLP. With respect to dialogue generation framework, the GAN-related researches could also be generally categorized as the GAN on generative-based and retrieval-based models.


As for sequence generation models, also regarded as sequential
decision making process in reinforcement learning, \newcite{DBLP:conf/aaai/YuZWY17} proposed seqGAN framework that bypasses the differentiation problem for discrete token generation by applying Monte Carlo roll-out policy, with recurrent neural network (RNN) as generator and binary classifier as discriminator. What follows are RankGAN~\cite{DBLP:conf/nips/LinLHSZ17} which treats the discrimination phase as a learning-to-rank optimization
problem as opposed to binary classification, dialogueGAN~\cite{DBLP:conf/emnlp/LiMSJRJ17} that adapts the GAN mechanism on a seq2seq model for dialogue generation scenario with its discriminator capable of identifying true query-response pairs from fake pairs, as well as DPGAN~\cite{DBLP:journals/corr/abs-1802-01345} that promotes response diversity by introducing an LM-based discriminator that overcomes the saturation problem for classifier-based discriminators. Nevertheless, even state-of-the-art generative approaches couldn't achieve comparable performance as retrieval-based approaches in terms of language fluency and diversity of response generation.

As for retrieval-based models, \newcite{DBLP:conf/sigir/WangYZGXWZZ17} proposed IRGAN framework that unifies both generative and discriminative ranking-based retrieval models through adversarial learning. While the generator learns the document relevance distribution and is able to generate (or select) unobserved documents that are difficult for discriminative ranker to rank correctly, the discriminator is trained to distinguish the good matching query-response pair from the bad. However effective IRGAN is, in a conversation scenario, a pure retrieval system would always be limited by the constructed query-response repository. The adversarial responses, observed or not, might not be suitable for novel queries after all, which is a common problem for retrieval-based conversation system that is beyond IRGAN's capability.

While previous GAN-related studies only focused on the improvement of either generative-based or retrieval-based single approach, our work in this paper could be categorized as a unified GAN framework of the aforementioned GAN mechanism on both retrieval model and sequence generation model of an ensemble, which is constructed with each of its modules getting involved in adversarial learning with different roles. While being most related to rankGAN and IRGAN, our work has the following differences:
\begin{itemize} 
\item[1)] RankGAN only trains a language model through point-wise ranking of independent human-written and synthetic sentences, while EnsembleGAN trains a generative seq2seq model (G$_1$) through pair-wise ranking (D) of ground-truth and negative responses, with both G$_1$ and D conditioned on the user's query, let alone the existence of another strong competitor G$_2$ providing negative adversarial samples.
\item[2)] While IRGAN allows for both generative and discriminative retrieval model to compete against each other, EnsembleGAN allows for both rankers G$_2$ and D to compete against each other as ensembles, with the constant involvement of response generation module G$_1$ included in a more delicate three-stage sampling strategy.
\item[3)] EnsembleGAN unifies both GAN mechanism with a shared overall learning objective among all generators and discriminator, enhancing an ensemble of generative and retrieval-based approaches towards better global optimal results.
\end{itemize}

\section{Preliminaries}
Before diving into details of our EnsembleGAN Framework, we first introduce the generation-based conversation model and the retrieval-based conversation model, which is the basis of our Ensemble model.

\paragraph{\textbf{Response Generation Model}}
An LM-based probabilistic conversation model usually employs the seq2seq encoder decoder framework, where in general the encoder learns the query representation and the decoder generates the response sequence token by token based on encoder output~\cite{NRM}. For an RNN-based seq2seq model with attention mechanism~\cite{DBLP:journals/corr/BahdanauCB14}, the generation probability of the current word $w_t$ of the response given query $q$ of length $T_q$ could be generally modeled as follows:
\begin{equation} 
\begin{split}
p(w_t|w_{t-1},\cdots, w_1, q) & = f_{\text{de}}(s_t,w_{t-1},c_t) \\
c_t & = f_\text{att}(s_t,\{h_i\}_{i=1}^{T_q}) \\
h_i & = f_{\text{en}}(w_i,h_{i-1})
\end{split}
\label{eq:seq2seq}
\end{equation}
where $f_{\text{en}}$ and $f_{\text{de}}$ are the recurrence functions. $h_i \in \mathbb{R}^{d_1}$ and $s_t\in \mathbb{R}^{d_2}$ represent the hidden state of the encoder and decoder, and $c_t$ the context vector obtained by attention mechanism based on $f_{\text{att}}$, which often takes the form of a weighted sum of $\{h_i\}_{i=1}^{T_q}$. The weight factor is generally computed as a similarity between $s_t$ and each $h_i\in \{h_i\}_{i=1}^{T_q}$, allowing the decoder to attend to different part of contexts at every decoding step. The cross entropy loss $\mathcal{L}_{ce}=-\sum y_t \log p(w_t)$ is often applied for the model training, with $y_t$ the ground-truth corresponding word.

\paragraph{\textbf{Response Ranking Model}}
Given a query $q$ and some candidate provided by a fast pre-retrieval module\footnote{We apply Lucene ({https://lucenenet.apache.org/}) to index all query-response pairs and use the built-in TF-IDF method to retrieve candidates, following~\newcite{DBLP:conf/ijcai/SongLNZZY18}.}, the ranking model learns to compute a relevance score between each candidate and the query $q$. Instead of the absolute relevance of individual responses (a.k.a, point-wise ranking), we train the model through pair-wise ranking, for a user's relative preference on a pair of documents is often more easily captured~\cite{DBLP:conf/sigir/WangYZGXWZZ17}. Hence, the probability of a response pair $\langle r_1,r_2\rangle$ with $r_1$ more relevant than $r_2$ (noted as $r_1 \succ r_2$) being correctly ranked can be estimated by the normalized distance of their matching relevance to $q$ as:
\begin{equation}
\label{eq:ranker module}
\begin{split} 
p(\langle r_1,r_2\rangle|q) & =\sigma(g(q,r_1)-g(q,r_2)) \\
& = \frac{\exp(g(q,r_1)-g(q,r_2))}{1+\exp(g(q,r_1)-g(q,r_2))}
\end{split}
\end{equation}
where $ \sigma $ is the sigmoid function, and $g(\cdot,\cdot)$ is the ranker's scoring function defined by any matching model. We train the ranker to rank the ground-truth response $r_{\text{pos}}$ higher than a sampled negative candidate $r_{\text{neg}}$, with the pair-wise ranking loss $L_{\text{rank}}$ defined as a hinge function~\cite{Herbrich2008Large}:
\vspace{-2mm}
\begin{equation} 
\small
\mathcal{L}_{\text{rank}} = \frac{1}{N}\sum_{i=1}^{N} \max(0, \delta +g(q,r_{\text{neg}})-g(q,r_{\text{pos}}))
\end{equation}
where $N$ is the number of $(q, \langle r_{\text{pos}},r_{\text{neg}}\rangle)$ training samples and $\delta$ denotes the margin allowing for a flexible decision boundary. While both the response generation and ranking model could be used alone as single model, they form an ensemble when the latter reranks both pre-retrieved candidates and generated responses and finally selects the response of the top ranking.

\section{EnsembleGAN Framework}


\begin{figure*}[htbp]
\centering
\resizebox{\linewidth}{!}{
    \includegraphics[width=7in]{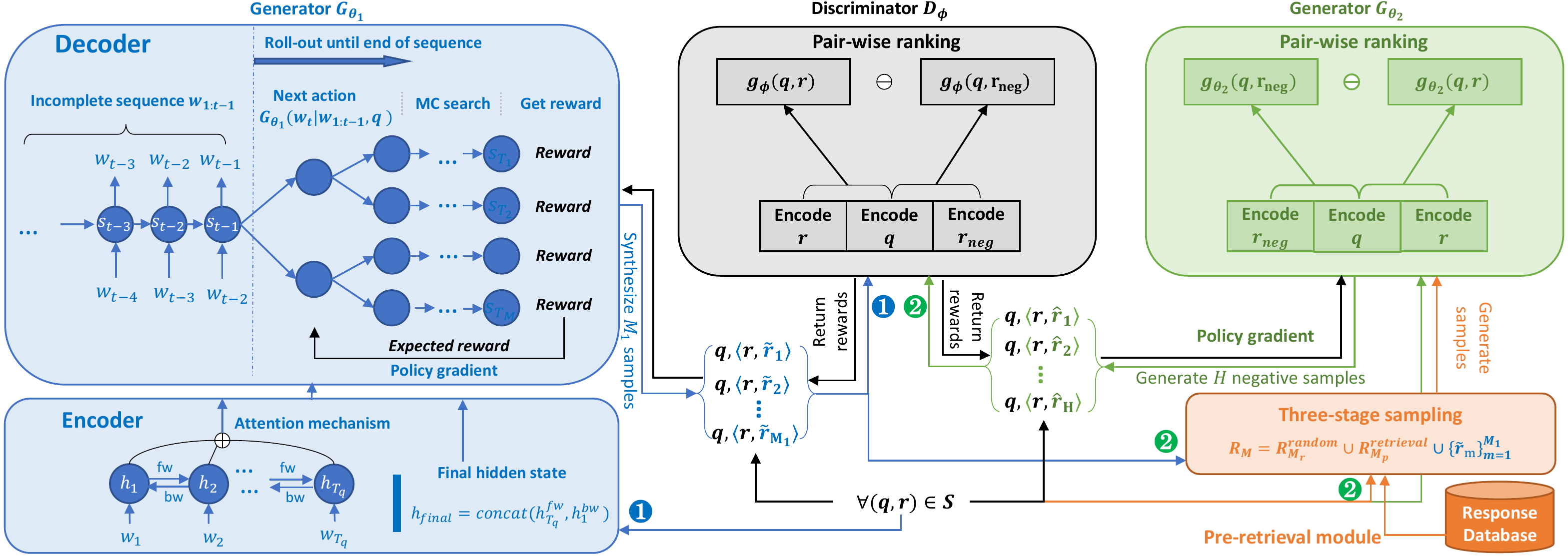}
}
\vspace{-6mm}
\caption{Illustration of EnsembleGAN Architecture (best viewed in color): generators $G_1$, $G_2$, discriminator $D$ as well as three-stage sampling strategy are represented by blue, green, grey and orange colored blocks respectively.\textcircled{1} and \textcircled{2} denote the training phase of $G_1$-steps and $G_2$-steps respectively, which is defined in algorithm \ref{alg:Framwork}.}
\vspace{-4mm}
\label{fig:1}
\end{figure*}

\subsection{Model Overview}
Figure~\ref{alg:Framwork} illustrates the overall architecture of our proposed EnsembleGAN framework. 
Given a set of user queries $Q=\{q_1,q_2,\cdots,q_N\}$, the original ensemble applies both its generation and pre-retrieval module to synthesize and retrieve response candidates $\{\tilde{r}_1, \cdots, \tilde{r}_{M_1}\}$ and $\{\hat{r}_1, \cdots, \hat{r}_H\}$ for each $q\in Q$, respectively. All candidates are ranked together based on the scoring function $g(q,r)$ of the ranking module. 

\paragraph{1) Generative seq2seq model} $G_{\theta_1}(\tilde{r}|q)$, which inherits from the generation module of the ensemble, is responsible for synthesizing response candidates $\{\tilde{r}_1, \cdots, \tilde{r}_{M_1}\}$ given query $q$, as depicted in \myref{eq:action value}, with the application of Monte Carlo (MC) roll-out policy. By combining with the ground-truth response $r$, we directly generate negative response pairs $\{\langle r,\tilde{r}_m\rangle\}^{M_1}_{m=1}$ aiming at receiving high ranking scores from discriminator, the process of which is also noted as $G_{\theta_1}(\langle r,\tilde{r}\rangle|q)$ for formulation coherence.

\paragraph{2) Generative ranking model} $G_{\theta_2}(\langle r,\hat{r}\rangle|q)$, which inherits from the response ranking model of the ensemble, learns to approximate the true relevance distribution over response pairs $p_{\text{true}}(\langle r,\hat{r}\rangle|q)$. Hence with the true response $r$, we generates highly competitive negative samples $\{\langle r,\hat{r}_h\rangle\}^H_{h=1}$ as specified in \myref{eq:Generative Ranker Model} so as to challenge the discriminator.

\paragraph{3) Discriminative ranking model} $D_\phi(\langle r,r_{\text{neg}}\rangle|q)$, which inherits from the same ranking model as $G_2$, endeavors however to distinguish the true response pairs from adversarial candidates provided by both generators ($G_1$ and $G_2$). After the adversarial training, all $G_1$, $G_2$ and $D$ could be used alone as single model, or we could also form an improved ensemble consisting of a generation model $G_1$ and a ranking model (either $G_2$ or $D$) as described previously.

\subsection{Adversarial Training for the Ensemble}
\subsubsection{Overall Objective} 
In our generative adversarial framework for the ensemble, both generators try to generate fake samples that get high ranking scores so as to fool the discriminator, the discriminator on the contrary is expected to distinguish the good samples from the bad by ranking more precisely as well as scoring down negative samples. We summarize the minimax game among generators $G_1$,$G_2$ and the discriminator $D$ with the objective function $\mathcal{L}$ as follows:
\begin{equation}
\small
\label{eq:overall loss}
\begin{split}
\mathcal{L} & = \min_{\theta_1,\theta_2}\max_{\phi}L(G_{\theta_1},G_{\theta_2},D_\phi) = \min_{\theta_1,\theta_2}\max_{\phi} (\mathcal{L}_1+\mathcal{L}_2)\\
\mathcal{L}_1 & =  \sum_{n=1}^N \displaystyle \mathbb{E}_{o\sim p_{\text{true}}(o|q_n)}[\log(D_\phi(o|q_n)] \\
\mathcal{L}_2 & =  \sum_{n=1}^N \displaystyle \mathbb{E}_{o'\sim G_{\theta_1,\theta_2}(o'|q_n)}[\log(1-D_\phi(o'|q_n))]
\end{split}
\end{equation}
where $\displaystyle \mathbb{E}$ denotes the mathematical expectation, $N$ the number of training samples, $o=\langle r,r_{\text{neg}}\rangle$ and  $o'=\langle r,r'_{\text{\text{neg}}}\rangle$ are the true and generated response pair, respectively.


\subsubsection{Optimizing Discriminative Ranker} 
As shown in \myref{eq:ranker module} previously, we design $D_\phi(\langle r,r_{\text{neg}}\rangle|q)=p_\phi(\langle r,r_{\text{neg}}\rangle|q)$ to evaluate the probability of a response pair $\langle r,r_{\text{neg}}\rangle$ being correctly ranked given query $q$. Combining the ground-truth responses with the fake ones generated by both current $G_1$ and $G_2$, the optimal parameters of $D_\phi$ are obtained as follows:
\vspace{-1mm}
\begin{equation} 
\small
\label{eq:train D}
\phi^{*}=\mathop{\mathrm{argmax}}_\phi (\mathcal{L}_1+\mathcal{L}_2)
\end{equation}
where $\mathcal{L}_1$ and $\mathcal{L}_2$ are defined in \myref{eq:overall loss}, such an optimization problem is usually solved by gradient descent as long as $D_\phi$ is differentiable with respect to $\phi$. When training generators, $D_\phi$ is used to provide reward of generated negative samples, which will be detailed later in this section.

\subsubsection{Optimizing Generative Seq2Seq} 
At the first stage, we enhance the generative seq2seq model $G_{\theta_1}$ through discriminative ranker $D_\phi$. When given a user query $q$, the generation of a sequence $\tilde{r}=\{w_0,w_1,...,w_T\}$ could be regarded as a series of decision making at $T$ time steps by policy $\pi=G_{\theta_1}(w_t|w_{1:t-1}, q)$ as defined in \myref{eq:seq2seq}. However, since $D_\phi$ only provides the reward for a complete sequence, the lack of intermediate reward for every time step leads to the ignorance of long term reward causing the model to be shortsighted. We hence apply MC roll-out policy~\cite{DBLP:conf/aaai/YuZWY17,DBLP:conf/nips/LinLHSZ17} to tackle with the problem, which repeatedly rolls out incomplete sequences until the end-of-sequence token so as to get an expected reward from $D_\phi$ for every time step. With the true response $r$, the expected end reward of a response pair $o'=\langle r,\tilde{r}\rangle$ is defined as follows:
\vspace{-0.5mm}
\begin{equation}
\small
\label{eq:J1}
\begin{split}
J_{\theta_1} (o'|q) & = \sum_{t=1}^T \displaystyle \mathbb{E}_{o'_m\sim G_{\theta_1}(o'_m|q)}[\log D_\phi(o'_m|q)|w_{1:t-1}] \\
& =\sum_{t=1}^T \sum_{w_t}G_{\theta_1}(w_t|w_{1:t-1}, q)Q_{D_\phi}^{G_{\theta_1}}(w_{1:t-1},w_t|q,\text{MC}_{m_1}^{\pi_r})
\end{split}
\vspace{-1mm}
\end{equation}
where $w_{1:t-1}$ is the current state with $t-1$ tokens already generated in $\tilde{r}$, $w_0$ the initial token. The response pair $o'_m=\langle r,w_{1:T_m}\rangle$, where $w_{1:T_m}$ is the completed $T_m$-length sequence rolled out from current $w_{1:t-1}$ according to $m_1$-time MC roll-out policy $\pi_r$ (noted as $\text{MC}_{m_1}^{\pi_r}$), resulting in the action-value function defined as follows:
\vspace{-1mm}
\begin{equation}
\small
\label{eq:action value}
\begin{split}
Q_{D_\phi}^{G_{\theta_1}}(w_{1:t-1},w_t|q,\text{MC}_{m_1}^{\pi_r})= 
\left\{
\begin{aligned}
& \frac{1}{m_1}\sum_{m=1}^{m_1}\log D_\phi(o'_m|q), & \text{for} \, t<T \\
& \log D_\phi(o'|q), & \text{for} \, t=T
\end{aligned}
\right.
\end{split}
\vspace{-1mm}
\end{equation}

Hence, the instant reward for time step $t$ is calculated as the average ranking scores from $D_\phi$ of all sampled response pairs $\{o'_m\}^{m_1}_{m=1}$ obtained by repeatedly rolling out $w_{1:t-1}$ for $m_1$ times based on $\text{MC}_{m_1}^{\pi_r}$. We note $M_1=m_1*T$ as the total number of generations for $\tilde{r}$ of length $T$. In contrast to the original rankGAN, both generator $G_{\theta_1}(w_t|w_{1:t-1}, q)$ and discriminator $D_\phi(o'|q)$ are conditioned on the given query $q$, which is a necessary adaptation in the case of dialogue generation. Note that such a configuration could also be referred to as conditionalGAN framework~\cite{DBLP:journals/corr/MirzaO14}.
\subsubsection{Optimizing Generative Ranker} 
The second stage involves the amelioration of generator $G_2$ through discriminator $D_\phi$, with the objective function defined as below:
\begin{equation}
\label{eq:J2}
J_{\theta_2|\theta_1}(q)= \displaystyle \mathbb{E}_{o'\sim G_{\theta_2|\theta_1}(o'|q)}[\log(1-D_\phi(o'|q))]
\end{equation}
where $\theta_2|\theta_1$ denotes that this second stage is actually based on the first stage discussed above, with $G_{\theta_1}$ fixed as a result. Inheriting from the same ranking model as $D_\phi$, we train $G_{\theta_2}$ to generate competitive negative response pairs that receive high ranking scores from $D_\phi$, where both ranking-based generative and discriminative models could get improved~\cite{DBLP:conf/sigir/WangYZGXWZZ17}. More precisely, when given a true $(q,r)$ pair and a scoring function $g_{\theta_2}$, the chance of $G_{\theta_2}$ selecting a negative sample $o'_h=(r,\hat{r}_h)$ according to the relevance distribution of response pairs $\{\langle r,\hat{r}_h\rangle|\hat{r}_h\succ r, \hat{r}_h\in R_M\}$ is defined by a softmax function as follows:
\vspace{-1mm}
\begin{equation} 
\small
\label{eq:Generative Ranker Model}
\begin{split}
G_{\theta_2}(o'_h|q) & = \frac{\exp(g_{\theta_2}(q,\hat{r}_h)-g_{\theta_2}(q,r))}{\sum_{\hat{r}_h\in R_M} \exp(g_{\theta_2}(q,\hat{r}_h)-g_{\theta_2}(q,r))}\\
& = \frac{\exp(g_{\theta_2}(q,\hat{r}_h))}{\sum_{\hat{r}_h\in R_M} \exp(g_{\theta_2}(q,\hat{r}_h))} = P_{\theta_2}(\hat{r}_h|q)
\end{split}
\end{equation}
where $R_M$ represents the M-sized candidate pool with ground-truth responses excluded. Despite other possible configurations as observed in \newcite{DBLP:conf/sigir/WangYZGXWZZ17}, we follow $G_{\theta_2}(o'_h|q)=P_{\theta_2}(\hat{r}_h|q)$ as directly being the relevance distribution of an individual response $\hat{r}_h$, not only for the simplicity, but for the coherence of both $G_{\theta_1}$ and $G_{\theta_2}$ being able to sample responses independently of the ground-truth response, as it's the real usage case after training.

The candidate pool $R_M$ is of importance for the capability of $G_{\theta_2}$ to sample $H$ unobserved as well as highly competitive responses. In addition to the random sampling strategy that generates $M_r$ random responses ($R_{M_r}^{\text{random}}$) from the database as the original IRGAN, we apply both the pre-retrieval module to retrieve $M_p$ candidates ($R_{M_p}^{\text{retrieval}}$) similar to ground-truth responses regardless of queries, and also $G_{\theta_1}$ to synthesize $M_1$ relevant responses, all of which are summarized as a three-stage sampling strategy:
\vspace{-1mm}
\begin{equation} 
\label{eq:sampling}
R_{M}(M_r,M_p,M_1)=R_{M_r}^{\text{random}}\cup R_{M_p}^{\text{retrieval}}\cup \{\tilde{r}_m\}^{M_1}_{m=1}
\vspace{-1mm}
\end{equation}

The design of $R_M$ not only compensates for the ineffectiveness of random sampling for generating competitive responses from a huge dialogue database in our case, it also enables the generator $G_2$ to work as an ensemble with the response generation model $G_1$, thus always considering the cooperation of both generative-based and retrieval-based approaches during adversarial learning.

\begin{algorithm}[!t]  
\small
    \caption{EnsembleGAN Minimax Game} 
    \label{alg:Framwork} 
    \begin{algorithmic}[1]
    \REQUIRE ~~\\
    Generators $G_{\theta_1}$,$G_{\theta_2}$, and discriminator $D_\phi$; \\
    Training data $\mathscr{D}_{\text{s2s}}$, $\mathscr{D}_{\text{rank}}$ and retrieval database $\mathscr{D}_{\text{ret}}$;\\
    Three-stage sampling approach $R_M$ as in \myref{eq:sampling}; \\
    $M_1$,$H$ the sampling size of $G_{\theta_1}$ and $G_{\theta_2}$ respectively. 
    \ENSURE ~~\\
    Ensemble of seq2seq model $G_{\theta_1}$ and ranker model $G_{\theta_2}$, $D_\phi$
    \STATE Initialize $G_{\theta_1}$, $D_\phi$ with random weights $\theta_1$,$\phi$; 
    \STATE Pretrain $G_{\theta_1}$, $D_\phi$ on $\mathscr{D}_{\text{s2s}}$, $\mathscr{D}_{\text{rank}}$ respectively
    \FOR{$G_1$-steps}
    \STATE $G_{\theta_1}(\cdot|q)$ generates $M_1$ samples for each $(q,r)\in \mathscr{D}_{\text{s2s}}$;
    \STATE Update $G_{\theta_1}$ via policy gradient defined in \myref{eq:train G};
    \ENDFOR
    \FOR{$G_2$-steps}
    \FOR{each $(q,r)\in \mathscr{D}_{\text{rank}}$}
    \STATE $G_{\theta_1}(\cdot|q)$ generates $M_1$ samples
    \STATE $G_{\theta_2}(\cdot,r|q)$ generates $H$ samples via $R_M$;
    \ENDFOR
    \STATE Update $G_{\theta_2}$ via policy gradient defined in \myref{eq:train G};
    \ENDFOR
    \FOR{$D$-steps}
    \STATE $G_{\theta_1}(\cdot|q)$ generates $M_1$ samples for each $(q,r)\in \mathscr{D}_{\text{rank}}$;
    \STATE $G_{\theta_2}(\cdot,r|q)$ generates $H$ samples via $R_M$ and combine with positive samples from $\mathscr{D}_{\text{rank}}$;
    \STATE Train discriminator $D_\phi$ according to \myref{eq:train D}
    \ENDFOR
\end{algorithmic}
\end{algorithm}

\subsubsection{Policy Gradient}
Following~\newcite{DBLP:conf/nips/SuttonMSM99}, we apply policy gradient to update generators' parameters through feedback of $D_\phi$ , for the sampling process of both generators are non-differential. Hence, with $D_\phi$ fixed, for each query $q$ with true-negative response pair $o'=(r,r'_{\text{neg}})$, the minimization of $\mathcal{L}$ in \myref{eq:overall loss} with respect to $\theta_1$,$\theta_2$ could be deduced as follows~\cite{DBLP:conf/nips/LinLHSZ17,DBLP:conf/sigir/WangYZGXWZZ17}:
\vspace{-1mm}
\begin{equation}
\small
\label{eq:train G}
\begin{split}
&\min_{\theta_1, \theta_2} \mathcal{L}= \max_{\theta_1}\sum_{n=1}^N \displaystyle \mathbb{E}_{o'\sim G_{\theta_1}}J_{\theta_1}(o'|q_n) - \max_{\theta_2|\theta_1}\sum_{n=1}^N J_{\theta_2|\theta_1}(q_n)\\
&\nabla_{\theta_1} J_{\theta_1}(o'|q_n) \simeq \sum_{t=1}^T \sum_{w_t}\nabla_{\theta_1} \log G_{\theta_1}(w_t|w_{1:t-1}, q_n)Q_{D_\phi}^{G_{\theta_1}} \\
&\nabla_{\theta_2} J_{\theta_2}(q_n) \simeq \frac{1}{H}\sum_{h=1}^H \nabla_{\theta_2} \log G_{\theta_2}(o'_h|q_n) \log D_\phi(o'_h|q_n)
\end{split}
\vspace{-1mm}
\end{equation}
where $J_{\theta_1}$, $J_{\theta_2}$ are defined in \myref{eq:J1} and \myref{eq:J2} respectively. $\nabla$ is the differential operator, $T$ the generated sequence length by $G_{\theta_1}$ and $H$ the negative sampling size of $G_{\theta_2}$.

\vspace{-2mm}
\subsubsection{Reward Setting}
Normally, we would consider that the reward $R\equiv \log D_\phi(r,r_{\text{neg}}|q)$. It's however problematic that the logarithm leads to instability of training~\cite{DBLP:conf/nips/GoodfellowPMXWOCB14}. We thus follow \newcite{DBLP:conf/sigir/WangYZGXWZZ17} with the advantage function of reward implementation defined as below:
\begin{equation}
\begin{split}
R & =2\cdot D_\phi(r,r_{\text{neg}}|q)-1 \\
& =2\cdot [\sigma(g_\phi(q,r)-g_\phi(q,r_{\text{neg}}))]-1
\end{split}
\end{equation}

\subsubsection{Overall Algorithm} We summarize the ensembleGAN algorithm in Algorithm~\ref{alg:Framwork}, where all the generators $G_1$, $G_2$ and discriminator $D$ are initialized by a pretrained ensemble, with $G_2$ and $D$ sharing the same parameter initialization.

Despite the very existence of Nash equilibrium between generator and discriminator for their minimax game, it remains an open problem of how they could be trained to achieve the desired convergence~\cite{DBLP:conf/nips/GoodfellowPMXWOCB14}. In our empirical study, we confirm that both the ranker  $D_\phi$ and generator $G_1$ are enhanced by ensembleGAN, while the ranker generator $G_2$ encounters a loss of performance after adversarial training, as also observed in~\newcite{DBLP:conf/sigir/WangYZGXWZZ17}.   

\section{Experiments}
In this section, we compare our EnsembleGAN with several representative GAN mechanism on a huge dialogue corpus. 
The goal of our experiments is to 1) evaluate the performance of our generation module and retrieval module for response generation and selection, and 2) evaluate the effectiveness of our proposed EnsembleGAN framework from the ensemble perspective.
\begin{table}[!t]
\centering
\caption{The Statistics of Mixed Short-Text Conversation Dataset. Resp. is response for short, \# Sent, \# Vocab and Avg\_L denote the number of sentences, vocabularies and the average sentence length, respectively.}
\vspace{-3mm}
\resizebox{0.48\textwidth}{!}{
    \begin{tabular}{c|l|c|c|c|c}
    \Xhline{1pt}
    \multicolumn{2}{c|}{\backslashbox{Features}{Dataset}}                     & Retrieval Pool & Ranking Set                                                & Generation Set & Test Set \\ \hline
    \multicolumn{2}{c|}{Corpus} & Weibo+Toutiao  & Weibo & Toutiao        & Toutiao  \\ \hline
    \multirow{3}{*}{Post} & \# Sent & 2,065,908 & 30,000  & 1,000,000  & 2,000    \\ \cline{2-6} 
    & \# Vocab       & 251,523 & 29,272 & 120,996 & 5,642 \\ \cline{2-6} 
    & Avg\_L & 11.4 & 13.1 & 9.3 & 10.1 \\ \hline
    \multirow{3}{*}{Resp.} & \# Sent    & 5,230,048      & 360,000 & 1,000,000      & 2,000    \\ \cline{2-6} 
     & \# Vocab & 628,254 & 28,000 & 121,763 & 4,544\\ \cline{2-6} 
    & Avg\_L & 8.7 & 9.8 & 7.1 & 7.7         \\ \hline
    Pair & \# Pairs & 6,000,000 & 360,000 & 1,000,000 & 2,000    \\ 
    \Xhline{1pt}
    \end{tabular}
}
\vspace{-4mm}
\label{tb:dataset}
\end{table}

\subsection{Dataset}
We conduct our experiments on a large mixed dialogue dataset crawled from online Chinese forum Weibo\footnote[1]{https://www.weibo.com/} and Toutiao\footnote[2]{https://www.toutiao.com/} containing millions of query-response pairs. For data pre-processing, we remove trivial responses like "wow" as well as the responses after first 30 ones for topic consistency following~\newcite{NRM}. We use Jieba\footnote[3]{https://github.com/fxsjy/jieba}, a common Chinese NLP tool, to perform Chinese word segmentation on all sentences. Each query and reply contain on average 10.2 tokens and 8.44 tokens, respectively. From the remaining query-response pairs, we randomly sampled 6,000,000 pairs as retrieval pool for the pre-retrieval module, 1,000,000 and 50,000 pairs for training and validating the sequence generation model, 360,000 and 2000 pairs for training and validating the ranking model (we apply three-stage sampling strategy to generate 11 negative samples for 30,000 true query-response pairs), and finally 2,000 pairs as test set for both models. We make sure that all test query-response pairs are excluded in training and validation sets. More detailed data statistics are summarized in \mytbref{tb:dataset}.

\subsection{Baselines}
We introduce baseline models and GAN competitors on three levels, namely the generation approach, the retrieval approach and the ensemble approach. We note GAN-G (D) for the generator (discriminator) of a GAN mechanism in this section. EnsembleGAN is represented by ensGAN for ease of demonstration.

\textbf{DialogueGAN.} We consider dialogueGAN~\cite{DBLP:conf/emnlp/LiMSJRJ17} as our GAN competitor for the generation part, with a seq2seq generator and a binary-classifier-based discriminator that is trained to distinguish the true query-response pairs from the fake ones. In order to eliminate structure biases for a fair comparison, we adopt the very same deep matching model structure as our ranking model (which will be detailed later) for its discriminator, instead of the hierarchical recurrent architecture applied by the original paper.

\textbf{DPGAN.} We consider diversity-promoting GAN (DPGAN)~\cite{DBLP:journals/corr/abs-1802-01345} as a second GAN competitor for the generation part, with a seq2seq generator and a language-model-based discriminator that is trained to assign higher probability (lower perplexity) for true responses than fake ones. The LM-based discriminator is consisted with a uni-directional LSTM~\cite{DBLP:journals/neco/HochreiterS97} as the original paper.

\textbf{RankGAN.} We consider RankGAN as another GAN competitor. The original RankGAN~\cite{DBLP:conf/nips/LinLHSZ17} is an unconditional language model that is unsuitable for dialogue generation scenario as discussed previously, we hence modify RankGAN to consist of a seq2seq generator and a pairwise discriminative ranker, which could be considered as ensGAN without getting generator G$_2$ involved. 

\textbf{IRGAN.} We also consider IRGAN~\cite{DBLP:conf/sigir/WangYZGXWZZ17} as a GAN competitor. Similarly, this could be considered as ensGAN without any involvement with seq2seq generator or the three-stage sampling strategy. All GAN mechanism are applied on exactly the same pre-trained generation or ranking model for a fair comparison, and we evaluate single component (generator or discriminator) as well as the derived ensemble (Generation + Ranking) for each GAN mechanism, resulting in various combinations which will be detailed later.

\textbf{Response Generation Models (S2SA).} We compare with the attention-based seq2seq model (S2SA)~\cite{DBLP:journals/corr/BahdanauCB14}, which has been widely adopted as a baseline model in recent studies~\cite{NRM,DBLP:conf/emnlp/LiMSJRJ17}. As a result, we have three derived adversarial sequence generators, namely the dialogueGAN-G, DPGAN-G, RankGAN-G that compete against ensGAN-G$_1$. Besides, We include mutual information enhanced seq2seq model (MMI-S2SA)~\cite{DBLP:conf/naacl/LiGBGD16} as another generative baseline method. 

\textbf{Pre-Retrieval Module (TF-IDF).} The pre-retrieval module, as the basis of retrieval approach, first calculates similarities among utterances (queries) based on simple TF-IDF scores and then retrieve the corresponding responses~\cite{DBLP:conf/ijcai/SongLNZZY18}. We report the Top-\{1,2\} responses, noted as TF-IDF-\{1,2\}, respectively.



\textbf{Response Ranking Models (Ranking).} The pure retrieval system is consisted with a pre-retrieval module and a ranking (matching) model, where the pre-retrieved candidates is reranked by the ranker, for which we apply state-of-the-art attentive conv-RNN model~\cite{DBLP:conf/kdd/WangJY17} for our ranker baseline. Therefore, we have 5 derived adversarial rankers based on the same original ranker, namely the RankGAN-D, IRGAN-G and IRGAN-D that compete against our ensGAN-G$_2$ and ensGAN-D. 

\textbf{Ensemble Models (Generation+Ranking).} Ensemble models are constructed with a generation model, a pre-retrieval module and a ranking model. When given a query, the generative model (e.g., S2SA, RankGAN-G and ensGAN-G$_1$) synthesizes candidate responses. Then the ranking model (e.g., conv-RNN, IRGAN-D, RankGAN-D and ensGAN-D) is required to rerank both pre-retrieved candidates and synthetic responses, and select the top one response in the end. Besides, following \newcite{DBLP:conf/ijcai/SongLNZZY18} and \newcite{DBLP:journals/corr/abs-1806-07042}, we also consider Multi-Seq2Seq + GBDT reranker and Prototype-Edit as two baseline ensemble models.

\subsection{Implementation Details} \label{sec:imple}
The seq2seq model is trained with a word embedding size of 300 for source and target vocabulary of 30,000 most frequent tokens of queries and responses in the generation training set, covering 97.47\% and 97.22\% tokens that appear in queries and responses respectively. The rest tokens are treated as "UNK" as unknown tokens. We set the hidden size of the encoder and decoder to 512. The adversarial sampling size $m_1=20$ during G$_1$ training steps.

The conv-RNN ranker is trained with 200-dimensional word embedding for a shared vocabulary of 40,000 tokens, covering 93.54\% words in the retrieval pool. The size of GRU is set to 200. The window size of the convolution kernel is set to (2, 3, 4, 8), with number of filters equal to (250, 200, 200, 150), following~\newcite{DBLP:conf/kdd/WangJY17}. We pretrain the ranker to rank the ground-truth response to the top from $k=11$ negative samples including 5 random samples, the top 5 pre-retrieved candidates and a synthetic one generated by seq2seq model. During adversarial training, the ranker generator $G_2$ generates $H=8$ negative samples from a candidate pool $R_M(100,10,10)$ according to the three-stage sampling strategy.

We use dropout of 0.2 for all models, and Adam optimizer~\cite{DBLP:journals/corr/KingmaB14} with a mini-batch of 50. The learning rate of S2SA and conv-RNN are respectively $0.0002$ and 0.001 during pre-training, $2\times 10^{-6}$ and $1\times 10^{-5}$ during adversarial learning. 

\begin{table*}[!t]
\centering
\caption{Overall performance of baselines and GAN competitors.  Ranking($\mathcal{M}$) means that candidate responses (generated by the pre-retrieval module) are re-ranked by the ranking model $\mathcal{M}$. Bold scores denote the highest score within each block. The RUBER scores for ground-truth are 0.815, 0.798 for RUBER$_A$ and RUBER$_G$, respectively.}  
\vspace{-2.5mm}
\resizebox{0.92\textwidth}{!}{
    \begin{tabular}{c|c|c|c|c|c|c|c|c|c|c} 
    \Xhline{1pt} 
    \multicolumn{2}{c|}{\multirow{2}{*}{\backslashbox{Modules}{Automatic Metrics}}} & \multicolumn{4}{c|}{Word Overlap} & \multicolumn{3}{c|}{Embedding Similarity} & \multicolumn{2}{c}{RUBER}  \\ \cline{3-11}
    \multicolumn{2}{c|}{} & BLEU$_1$ & BLEU$_2$ & BLEU$_3$ & BLEU$_4$ & EA & GM & VE & RUBER$_A$ & RUBER$_G$  \\ \hline
    \multirow{6}{*}{Generation} & S2SA & 7.334  & 2.384  & \textbf{0.987}  & 0.340 & 0.503 & 0.154 & 0.332 &0.550 & 0.500  \\ 
    & MMI-S2SA &  8.468 & 2.464 & 0.956 & 0.404 & 0.526 & 0.149 & 0.342 & 0.557 & 0.521 \\
    & DialogGAN-G &  9.465 & 2.483 & 0.912 & 0.349& 0.533&0.161&0.344 &0.560 & 0.533 \\ 
    & DPGAN-G & 8.578  & 2.474 & 0.922 & 0.385 &  0.535 &\textbf{0.165} &0.345 & 0.588 & 0.557 \\ 
    & RankGAN-G & \textbf{10.033}  & \textbf{2.545} & 0.967  & \textbf{0.436} & \textbf{0.560} & 0.145 & 0.343 & \textbf{0.602} & 0.580    \\ 
    & ensGAN-G1 & 9.530  & 2.487  & 0.872  & 0.352 & 0.531 & 0.163 & \textbf{0.347} & 0.598 & \textbf{0.584}  \\   
    \hline
    \multirow{8}{*}{Retrieval}  & TF-IDF-1 (pre-retrieval) & 7.026  & 2.175  & 0.928  & 0.460 & 0.537 & 0.152 & 0.337 & 0.541 & 0.486   \\ 
    & TF-IDF-2 (pre-retrieval) & 7.120  & 2.108  & 0.990  & \textbf{0.581}  & 0.538 & 0.153 & 0.338 & 0.539 & 0.499  \\ 
    & Ranking (conv-RNN)  & 7.242  & 2.213  & 0.933  & 0.488  & 0.543 & 0.151 & 0.339 & 0.558 & 0.519    \\ 
    & Ranking (RankGAN-D) & 7.441  & 2.194  & 0.945  & 0.490   & 0.547 & 0.152 & 0.341 & 0.571 & 0.535  \\ 
    & Ranking (IRGAN-G)          & 7.225  & 2.166  & 0.867  & 0.409  & 0.540 & 0.152 & 0.337 & 0.560 & 0.518  \\ \ 
    & Ranking (IRGAN-D) & 7.451  & \textbf{2.362}  & \textbf{1.012}  & 0.528 & \textbf{0.553} & \textbf{0.156} & \textbf{0.343} & 0.573 & \textbf{0.542} \\ 
    & Ranking (ensGAN-G$_2$) & 7.057  & 2.129  & 0.897  & 0.460  & 0.539 & 0.150 & 0.338 & 0.549 & 0.516 \\ 
    & Ranking (ensGAN-D) & \textbf{7.452}  & 2.320  & 1.004  & 0.527& 0.548 & 0.153 & 0.341 & \textbf{0.579} & 0.539  \\ \hline
    \multirow{10}{*}{Ensemble} &  Multi-Seq2Seq + GBDT~\cite{DBLP:conf/ijcai/SongLNZZY18} & 7.542  & 2.173  & 0.993  & 0.569 & 0.540 & 0.152 & 0.338 & 0.592 & 0.568 \\ 
    &  Prototype-Edit~\cite{DBLP:journals/corr/abs-1806-07042} & 7.926  & 2.334  & 1.120  & 0.571 & 0.557 & 0.164 & 0.346 & 0.610 & 0.587   \\ 
    & S2SA + conv-RNN  & 7.630  & 2.299  & 1.125  & 0.555 & 0.544 & 0.153 & 0.341 & 0.564 & 0.535 \\ 
    & RankGAN-G + conv-RNN    & 7.755  & 2.275 & 0.889 & 0.432 & 0.549 & 0.150 & 0.340     &0.572 & 0.543   \\ 
    & ensGAN-G$_1$ + conv-RNN     & 7.570  & 2.168 & 0.871 & 0.423 & 0.544 & 0.156 & 0.339 &0.568 & 0.540  \\ 
    & RankGAN-G + IRGAN-D & 8.827 & 2.693 & 1.234 & 0.716 & 0.560 & 0.152 & 0.348 & 0.608 & 0.577 \\
    & S2SA + IRGAN-D       & 8.375  & 2.850  & 1.232  & 0.637 & 0.558 & 0.162 & 0.347& 0.600 & 0.573  \\ 
    & S2SA + ensGAN-D   & 8.535  & 2.749  & \textbf{1.297}  & 0.715 & 0.547 & 0.159 & 0.345  & 0.595 & 0.569 \\ 
    & RankGAN-G + RankGAN-D      & 8.715  & 2.501  & 1.075  & 0.580 & \textbf{0.561} & 0.154 & 0.347& 0.615 & 0.591  \\ 
    & ensGAN-G$_1$ + ensGAN-D & \textbf{9.339}  & \textbf{2.876}  & 1.277  & \textbf{0.763}  & 0.559 & \textbf{0.178} & \textbf{0.352}  & \textbf{0.621} & \textbf{0.605} \\

    \Xhline{1pt}
    \end{tabular}
}
\vspace{-2.5mm}
\label{tb:overall result}
\end{table*}
\subsection{Evaluation Metrics}
We adopt multiple automatic evaluation criteria as well as human evaluation for a comprehensive comparison. 

\textbf{BLEU.} BLEU~\cite{papineni2002bleu} evaluates the word-overlap between the proposed and the ground-truth responses. Typically, we use BLEU$_{n}$ ($n=1,2,3,4$) to calculate their n-grams-overlap, where BLEU$_n$ denotes the BLEU score considering n-grams of length $n$.

\textbf{Embedding-based metrics (EA, GM, VE).} Following \newcite{DBLP:conf/emnlp/LiuLSNCP16}, we alternatively apply three heuristics to measure the similarity between the proposed and ground-truth response based on pre-trained word embeddings\footnote[1]{We apply pre-trained Chinese word embedding which is available at https://github.com/Embedding/Chinese-Word-Vectors.}, including Embedding Average (EA), Greedy Matching (GM), and Vector Extrema (VE).

\textbf{Semantic Relevance (RUBER$_A$, RUBER$_G$).} Together with the embedding similarity, \newcite{DBLP:conf/aaai/TaoMZY18}  evaluates the semantic relatedness between a response and its query based on neural matching models. Following the original paper, we report the arithmetic and geometric mean of embedding similarity and semantic relatedness, denoted as RUBER$_A$ and RUBER$_G$, respectively.


\textbf{Retrieval Precision (P@1).} We evaluate pure ranking-based retrieval systems by precision at position 1 (P@1), which calculates the ratio of relevant responses (in our case, the ground-truth response) within top-$1$ reranked responses. 

\textbf{Human evaluation.} We also conduct human evaluations for generation and ensemble models since automatic metrics might not be consistent with human annotations~\cite{DBLP:conf/emnlp/LiuLSNCP16,DBLP:conf/aaai/TaoMZY18}. Following previous studies~\cite{DBLP:journals/corr/XingWWLHZM16,DBLP:conf/aaai/TaoMZY18, DBLP:conf/ijcai/SongLNZZY18}, we invited 3 well educated volunteers to judge the quality of 100 randomly generated responses by different models\footnote{Due to numerous generation + ranking possibilities and space limitations, we only asked annotators to evaluate representative models with high automatic metric scores.}, based on the following criteria: a score of $0$ indicates a bad response that is either dis-fluent or semantically irrelevant; $+1$ means a relevant but universal response;  $+2$ indicates a fluent, relevant and informative response. We report the proportion of each score ($0,+1,+2$) for each model.  Fleiss' kappa~\cite{Fleiss1971Measuring} scores are also reported.

\subsection{Results and Analysis}
\subsubsection{Overall Performance} Our evaluation is divided into three parts, namely the evaluation for pure generation module, pure retrieval module and the ensemble. \mytbref{tb:overall result} summarizes the general dialogue generation performance including various automatic metrics of word overlap, embedding similarity and semantic relevance. Figure~\ref{fig:precision} shows the P@1 scores for different retrieval systems as well as the study of contribution of two modules that consist of an ensemble, together with \mytbref{tb:result} of human evaluation results for representative models.
The human agreement is validated by the Kappa with a value range of 0.4 to 0.6 indicating ``moderate agreement" among annotators. A higher value denotes a higher degree of agreement, such as 0.65 for S2SA which is probably because it generates more dis-fluent or irrelevant responses that are easy to recognize. We first make several observations as follows:
\begin{table}[t]
\small
\centering
\caption{Results of human evaluation for generation and ensemble models.  ``Kappa" means Fleiss' kappa.}
\vspace{-2mm}
\resizebox{0.45\textwidth}{!}{
\begin{tabular}{l|c|c|c|c} 
    \Xhline{1pt}
    {\backslashbox{Model}{Score}} & +2  & +1  & 0 & Kappa  \\ \hline
    S2SA & 0.12 & 0.40 & 0.48 & 0.65 \\ 
    ensGAN-G$_1$ & 0.14 & 0.49 & 0.36 & 0.55\\ 
    RankGAN-G & 0.16 & 0.39 & 0.45 & 0.43\\ \hline 
    S2SA + conv-RNN & 0.21 & 0.33 & 0.47 & 0.52\\ 
    S2SA + IRGAN-D & 0.22 & 0.35 & 0.43 & 0.47\\ 
    S2SA + ensGAN-D & 0.25 & 0.35 & 0.40 & 0.49  \\ 
    RankGAN-G + RankGAN-D & 0.28 & 0.35 & 0.36 & 0.46\\ 
    RankGAN-G + IRGAN-D & 0.30 & 0.36 & 0.35 & 0.53\\ 
    ensGAN-G$_1$ + ensGAN-D & 0.37 & 0.38 & 0.26 & 0.45\\ 
    \Xhline{1pt}
\end{tabular}
}
\vspace{-4mm}
\label{tb:result}
\end{table}

\begin{description}[style=unboxed,leftmargin=0cm,font=\normalfont\space]
\item [1)] As for generation module, we first notice that GAN-enhanced seq2seq models achieve plausible improvement on most evaluation metrics, outperforming S2SA and MMI-S2SA baselines. Both RankGAN-G and ensGAN-G$_1$ aim at synthesizing responses that approximate true responses with higher ranking scores, which is demonstrated by the obvious gain of their contribution ratios for ensembles shown in Figure~\ref{fig:contribution}, with more than 40\% contribution for both RankGAN ensemble (E$_{Rank}$) and ensGAN ensemble (E$_{ens}$). Their comparable enhancement to RUBER scores indicates better generations in terms of the semantic relevance. Despite the outperforming word overlap and embedding average of RankGAN-G, ensGAN-G$_1$ is not only better at improving the GM and VE metrics, indicating more generation of key words with important information that are semantically similar to those in the ground-truth~\cite{DBLP:conf/emnlp/LiuLSNCP16}, but it's also capable of generating more satisfying responses with fewer 0 human scores according to \mytbref{tb:result}. 

\begin{figure}[t!]
  \centering
  \subfigure[] { \label{fig:precisionAt1}
    \includegraphics[height=0.445\columnwidth]{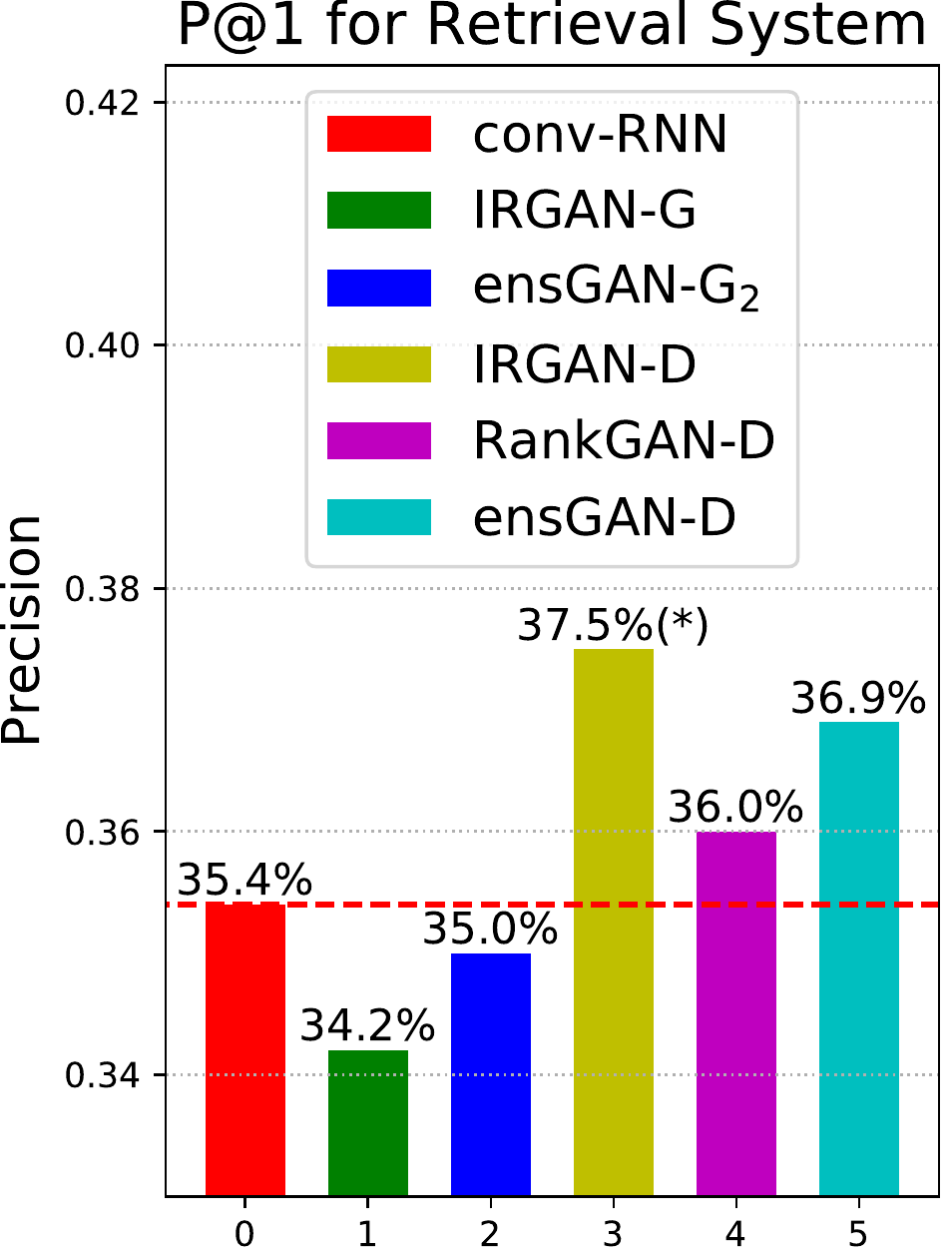}
  }
  \subfigure[] { \label{fig:contribution}
    \includegraphics[height=0.445\columnwidth]{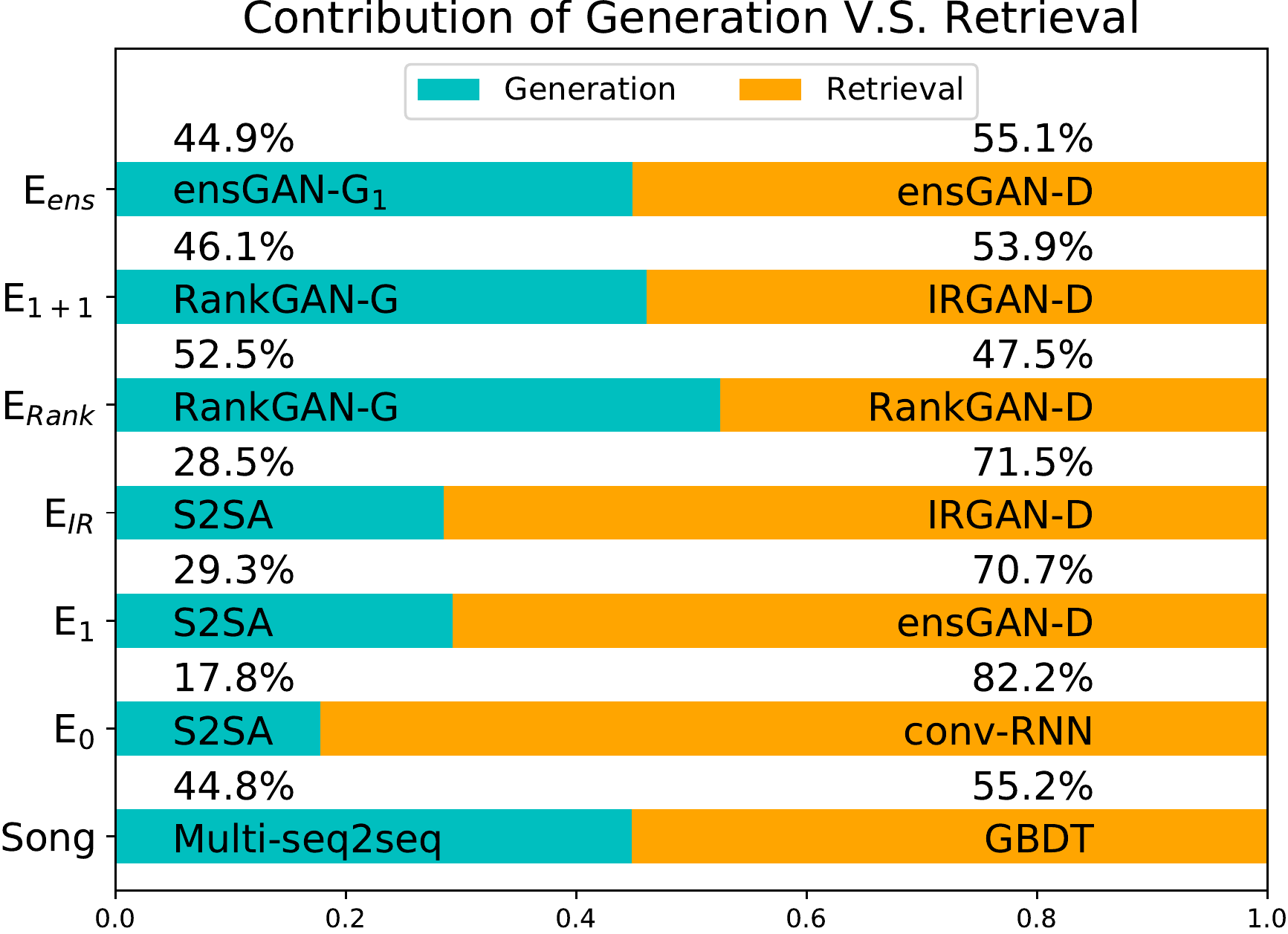}
  }
  \vspace{-5mm}
  \caption{(a) P@1 scores for various ranker-based retrieval systems. * denotes significant precision improvement (compared with conv-RNN) according to the Wilcoxon signed-rank test; and (b) The final response contribution of generation and ranking modules for ensembles.}
  \vspace{-5mm}
  \label{fig:precision}
\end{figure}

\item [2)]  As for retrieval methods, we see that they often achieve advantageous higher order BLEU scores (e.g., BLEU$_{3}$ and BLEU$_{4}$) than generative approaches, since generating responses of better language fluency (hence higher n-gram overlaps to some extent) is undoubtedly their strong points. They are however inferior to generative methods in terms of RUBER scores, for the latter are generally better at generating more tailored responses of high semantic relatedness~\cite{DBLP:conf/ijcai/SongLNZZY18}, similar results are also obtained by \newcite{DBLP:conf/aaai/TaoMZY18}. Together with P@1 scores in Figure~\ref{fig:precisionAt1}, all discriminative rankers of GAN approaches (IRGAN-D, RankGAN-D, ensGAN-D) are generally ameliorated on various aspects, with generative rankers (IRGAN-G, ensGAN-G$_2$) somehow deteriorated, which is also confirmed by~\newcite{DBLP:conf/sigir/WangYZGXWZZ17}. Similarly, one possible explanation might be the sparsity of the positive response distribution compared with negative ones during training, making it hard for a generative ranker to get positive feedbacks from discriminator. Without any generation module, IRGAN outperforms others on enhancing a pure retrieval system, notably achieving the highest P@1 score. On the other hand however, the P@1 score for all methods remains low compared with common QA task~\cite{DBLP:conf/sigir/WangYZGXWZZ17,DBLP:conf/kdd/WangJY17}, which might be explained by a more complicated and chaotic nature of STC dataset~\cite{DBLP:conf/emnlp/WangLLC13}.

\item [3)]  As for the ensembles, they commonly outperform previous single approaches, for the scores in the third block (Ensemble) are generally better than the first two blocks (Generation and Retrieval), which is especially true for those GAN-enhanced ensembles. Among various combinations of generation + ranking, the ensGAN ensemble (ensGAN-G$_1$ + ensGAN-D) outperforms both IRGAN (S2SA + IRGAN-D) and RankGAN (RankGAN-G + RankGAN-D) ensembles with the largest gain on almost all metrics, as well as achieving the most +2 and the fewest 0 scores for human judgement. While RankGAN and IRGAN bring specific enhancement to the generative and retrieval module respectively, the ensGAN improves the whole ensemble by allowing each its module to compete against each other, which might be regarded as seeking for a global optimum compared with other GAN that searches for local optimum of a single approach. While the ensGAN-G$_1$ generation module accounts more for the ensemble's final selection, the ensGAN-D learns to rank (select) responses featuring advantages of both generative and retrieval approach as expected, with the help of another strong negative sampler G$_2$ during adversarial training.
\end{description}
\begin{figure}[t!]
\centering
\resizebox{\linewidth}{!}{
    \includegraphics[width=10in,height=2.6in]{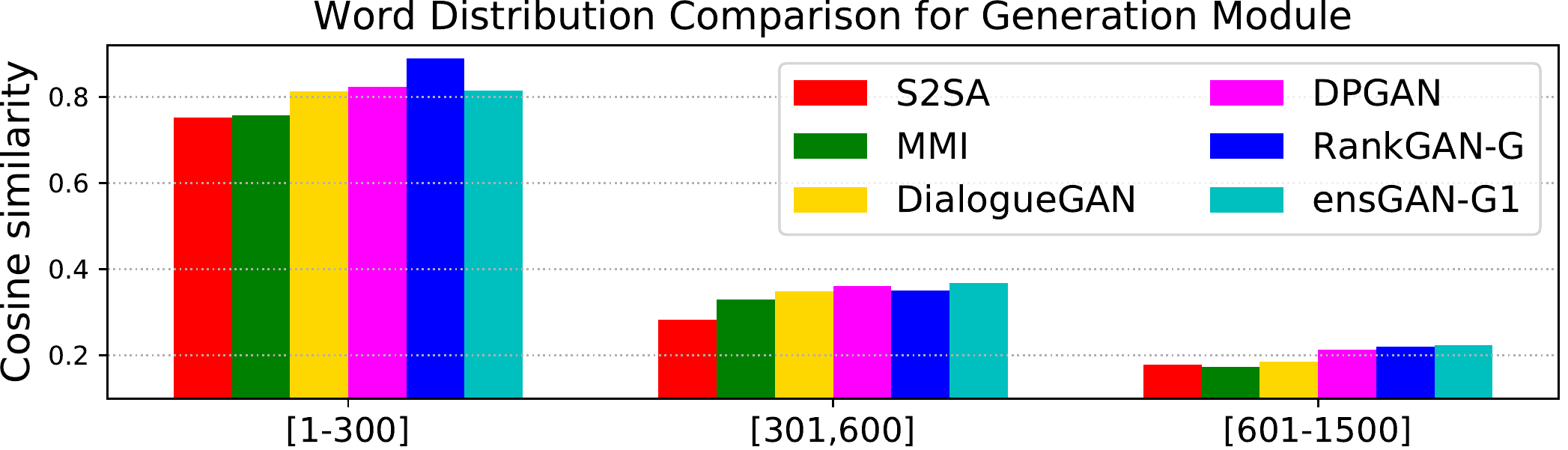}
}
\vspace{-6mm}
\caption{Cosine similarity between ground-truth and synthetic word distribution by various generative models on test data for different word frequency level (e.g., top 300 frequent words). EnsGAN achieves satisfying performance especially when considering words of lower frequency.}
\vspace{-5mm}
\label{fig:2}
\end{figure}

\vspace{-0.5mm}
\subsubsection{Discussion}
In addition to previous observations, we'd also like to provide further insights of the EnsembleGAN framework on several interesting aspects in this section.

\vspace{-0.5mm}
\paragraph{\textbf{Ranking versus LM versus Binary-Classification.}} As for the amelioration of generative seq2seq model, while DialogueGAN uses a binary classifier as the discriminator, DPGAN utilizes an LM-based discriminator, and both RankGAN and ensGAN apply ranking-based discriminator. As a result, the superiority of adversarial ranking over binary-classification is not only observed in our experiment, but confirmed in~\cite{DBLP:conf/nips/LinLHSZ17} as well. The LM-based discriminator (DPGAN-D) on the other hand, by addressing the saturation issue of binary classification~\cite{DBLP:journals/corr/abs-1802-01345}, brings comparable improvement as adversarial ranking, all of which help generate responses of higher quality as observed previously, as well as achieving better cosine similarity of word distributions (Figure~\ref{fig:2}). In particular, we apply the adversarial ranking in our work for it's the very bridge that connects the adversarial training of both generative-based and retrieval-based methods in the EnsembleGAN framework.

\begin{figure}[t!]
  \centering
  \subfigure[] { \label{fig:Pcurve}
    \includegraphics[height=0.43\columnwidth]{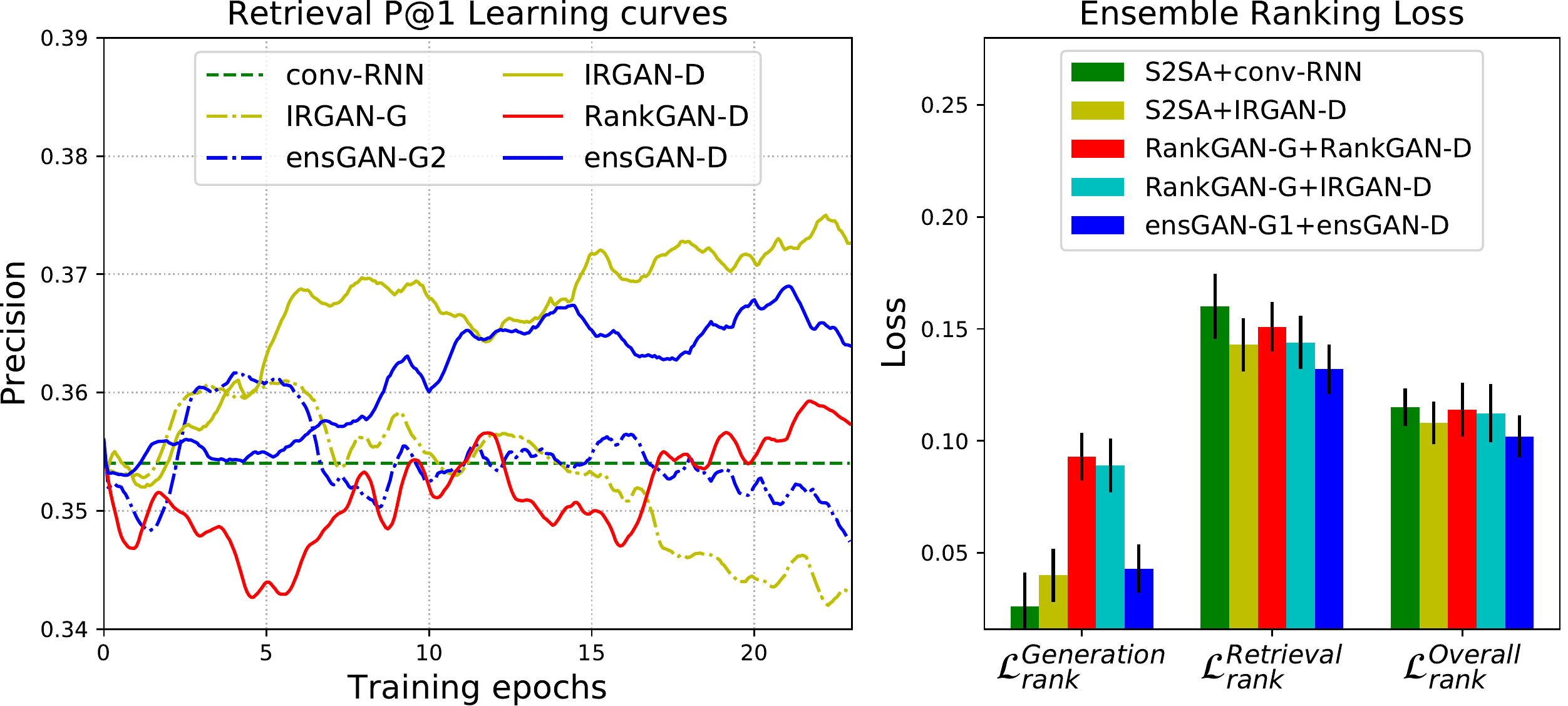}
  }
  \subfigure[] { \label{fig:Rankloss}
    \includegraphics[height=0.43\columnwidth]{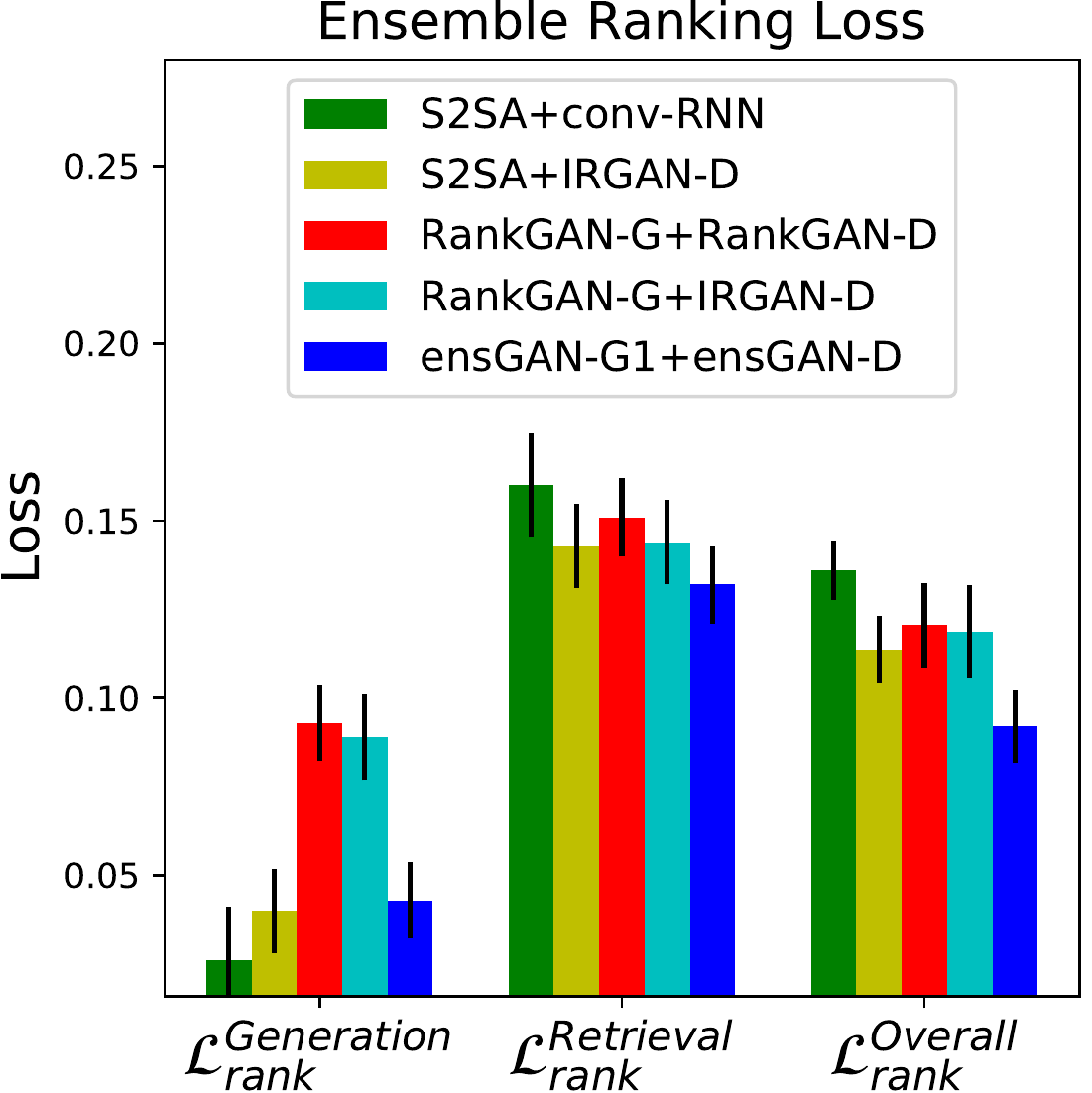}
  }
  \vspace{-5mm}
  \caption{(a) Ranker P@1 learning curves; and (b) Error bars of mean and standard deviation of ranking loss for different modules of ensembles. Results are calculated on test set.}
  \vspace{-5mm}
  \label{fig:loss}
\end{figure}

\vspace{-0.5mm}
\paragraph{\textbf{1+1$\neq$2 for Ensemble Approach.}} Although it's unreasonable to exhaustively study all possible generation + ranking combinations, it's however interesting to directly combine the seemingly best two modules of their separate worlds, namely the RankGAN-G + IRGAN-D, to see how such an ensemble performs compared with ensGAN. Apart from the overall results in \mytbref{tb:overall result} which already indicate that these two "best winners" do not get along as well as ensGAN-G$_1$ + ensGAN-D to some extent, a further evidence lies in the analysis on the ranking module shown in Figure~\ref{fig:loss}. On one hand, the P@1 adversarial learning curves (Figure~\ref{fig:Pcurve}) show that the IRGAN is better at enhancing a pure retrieval system, while RankGAN-D encounters a higher oscillation which is probably due to its concentration on ranking the synthetic responses to the top, making its P@1 pure retrieval performance unpredictable. On the other hand, the ensemble of ensGAN-G$_1$ + ensGAN-D turns out to be clearly advantageous in terms of the ranking loss (Figure~\ref{fig:Rankloss}) defined in \myref{eq:ranker module} among ensemble approaches. More specifically, we calculate the module-wise ranking loss for the final chosen responses (considered as $r_{\text{neg}}$) from the generation ($\mathcal{L}_{rank}^{Generation}$) or the pre-retrieval module ($\mathcal{L}_{rank}^{Retrieval}$), the overall ranking loss ($\mathcal{L}_{rank}^{Overall}$) is computed as the weighted sum of the two losses based on the module contribution. We see that ensGAN-D generally achieves the lowest ranking loss with moderate variance, which clearly demonstrates that EnsembleGAN is indeed more inclined towards global optimum without unilaterally enhancing a single module and thus is more adapted for an ensemble of multiple modules, especially when we note that the direct combination of the two "best winners" RankGAN-G + IRGAN-D does not result in the lowest overall ranking loss (not even close).


\vspace{-0.5mm}
\paragraph{\textbf{The Merits of the Ranking Module.}} In addition, we find that there also exists a clear performance gap among the ensembles themselves. As shown in \mytbref{tb:overall result}, the combinations of original S2SA + GAN-enhanced rankers generally bring better ameliorations compared with the combinations of GAN-enhanced S2SA + original conv-RNN, suggesting the very importance of a re-ranker for a dialogue ensemble, which is reasonable because all candidates have to be reranked by this final decision maker. Hence, despite the trend on the amelioration of generative approaches, it's also plausible to concentrate on the research of retrieval or ensemble methods so as to improve the open domain human-computer conversation.

\begin{table}[t!]
\centering
\caption{Response generation case study. The final decision of rankers are marked by $\surd$. White and gray cells denote valid and inaccessible candidates for a ranker when combined with its corresponding generative module as an ensemble. The original ranker is noted $D_{\text{O}}$, and $D_{\text{IR}}$, $D_{\text{R}}$, $D_{\text{E}}$, $D_{\text{T}}$ denote IRGAN-D, RankGAN-D, ensGAN-D and GBDT respectively.}
\vspace{-2mm}
\resizebox{0.48\textwidth}{!}{
\begin{tabular}{rl|c|c|c|c|c} 
\Xhline{1pt} 
& \textbf{Response generation cases} & $D_{\text{O}}$ & $D_{\text{IR}}$ & $D_{\text{R}}$ & $D_{\text{E}}$ &$D_{\text{T}}$ \\\hline \hline
$q$: \!\!&\!\!I can't play sniper. & \cellcolor[gray]{0.93} & \cellcolor[gray]{0.93} & \cellcolor[gray]{0.93} & \cellcolor[gray]{0.93} &\cellcolor[gray]{0.93}\\\cline{1-2}
$r$: \!\!&\!\!You're messing up with me.& \cellcolor[gray]{0.93} & \cellcolor[gray]{0.93} & \cellcolor[gray]{0.93} & \cellcolor[gray]{0.93} &\cellcolor[gray]{0.93}\\\hline
TF-IDF-1: \!\!&\!\!Really can't play this.  & $\surd$ & $\surd$ &&&\\ \hline
TF-IDF-2: \!\!&\!\!I really can't play games.  & &&& &$\surd$\\\hline
S2SA: \!\!&\!\!Have you ever played? & && \cellcolor[gray]{0.93} & \cellcolor[gray]{0.93}&\cellcolor[gray]{0.93}\\\cline{1-4}
MMI-S2SA: \!\!&\!\!Right, I've been playing for a year. & \cellcolor[gray]{0.93} & \cellcolor[gray]{0.93} & \cellcolor[gray]{0.93} & \cellcolor[gray]{0.93} &\cellcolor[gray]{0.93}\\\cline{1-2}
DialogueGAN: \!\!&\!\!I've played once, I don't know. & \cellcolor[gray]{0.93} & \cellcolor[gray]{0.93} & \cellcolor[gray]{0.93} & \cellcolor[gray]{0.93} &\cellcolor[gray]{0.93}\\\cline{1-2}
DPGAN: \!\!&\!\!They're my men, what're you thinking? & \cellcolor[gray]{0.93} & \cellcolor[gray]{0.93} & \cellcolor[gray]{0.93} & \cellcolor[gray]{0.93} &\cellcolor[gray]{0.93}\\\cline{1-2}\cline{5-5}
RankGAN-G: \!\!&\!\!You'll know once you've played. &\cellcolor[gray]{0.93} &\cellcolor[gray]{0.93}& $\surd$ &\cellcolor[gray]{0.93} &\cellcolor[gray]{0.93}\\\cline{1-2}\cline{5-6}
ensGAN-G1: \!\!&\!\!What I played was real. &\cellcolor[gray]{0.93} &\cellcolor[gray]{0.93}&\cellcolor[gray]{0.93}& $\surd$ &\cellcolor[gray]{0.93}\\\cline{1-2}\cline{6-7}
Multi-Seq2Seq: \!\!&\!\! I've played this game, and you? & \cellcolor[gray]{0.93} & \cellcolor[gray]{0.93} & \cellcolor[gray]{0.93} & \cellcolor[gray]{0.93} &\\ \hline\hline
$q$: \!\!&\!\!Looking for the title of this song. & \cellcolor[gray]{0.93} & \cellcolor[gray]{0.93} & \cellcolor[gray]{0.93} & \cellcolor[gray]{0.93} &\cellcolor[gray]{0.93}\\ \cline{1-2}
$r$: \!\!&\!\!It's the theme song from Rudy.& \cellcolor[gray]{0.93} & \cellcolor[gray]{0.93} & \cellcolor[gray]{0.93} & \cellcolor[gray]{0.93}&\cellcolor[gray]{0.93}\\ \hline
TF-IDF-1: \!\!&\!\!It's Faith. & $\surd$ & $\surd$ && &\\ \hline
TF-IDF-2: \!\!&\!\!Faith by Xinzhe Zhang. & &&& $\surd$ &$\surd$\\ \hline
S2SA: \!\!&\!\!UNK.  & & & \cellcolor[gray]{0.93} & \cellcolor[gray]{0.93} &\cellcolor[gray]{0.93}\\\cline{1-4}
MMI-S2SA: \!\!&\!\!I'd like to know where you're from. & \cellcolor[gray]{0.93} & \cellcolor[gray]{0.93} & \cellcolor[gray]{0.93} & \cellcolor[gray]{0.93} &\cellcolor[gray]{0.93}\\\cline{1-2}
DialogueGAN: \!\!&\!\!UNK. & \cellcolor[gray]{0.93} & \cellcolor[gray]{0.93} & \cellcolor[gray]{0.93} & \cellcolor[gray]{0.93} &\cellcolor[gray]{0.93}\\\cline{1-2}
DPGAN: \!\!&\!\!Yeah! & \cellcolor[gray]{0.93} & \cellcolor[gray]{0.93} & \cellcolor[gray]{0.93} & \cellcolor[gray]{0.93} &\cellcolor[gray]{0.93}\\\cline{1-2}\cline{5-5}
RankGAN-G: \!\!&\!\!UNK by UNK.  &\cellcolor[gray]{0.93} &\cellcolor[gray]{0.93}& $\surd$ &\cellcolor[gray]{0.93} &\cellcolor[gray]{0.93}\\\cline{1-2}\cline{5-6}
ensGAN-G1: \!\!&\!\!Thanks for your love. &\cellcolor[gray]{0.93} &\cellcolor[gray]{0.93}&\cellcolor[gray]{0.93}&  &\cellcolor[gray]{0.93}\\\cline{1-2}\cline{6-7}
Multi-Seq2Seq: \!\!&\!\! UNK. & \cellcolor[gray]{0.93} & \cellcolor[gray]{0.93} & \cellcolor[gray]{0.93} & \cellcolor[gray]{0.93} &\\ \hline\hline
$q$: \!\!&\!\!You're pretty, with or without fringe. & \cellcolor[gray]{0.93} & \cellcolor[gray]{0.93} & \cellcolor[gray]{0.93} & \cellcolor[gray]{0.93} &\cellcolor[gray]{0.93}\\ \cline{1-2}
$r$: \!\!&\!\!Well, I've got a big face actually.& \cellcolor[gray]{0.93} & \cellcolor[gray]{0.93} & \cellcolor[gray]{0.93} & \cellcolor[gray]{0.93}&\cellcolor[gray]{0.93}\\ \hline
TF-IDF-1: \!\!&\!\!Hahahaha. & &$\surd$ && &\\ \hline
TF-IDF-2: \!\!&\!\!UNK. &$\surd$ &&& &\\ \hline
S2SA: \!\!&\!\!Thanks.  &  &  & \cellcolor[gray]{0.93} & \cellcolor[gray]{0.93} &\cellcolor[gray]{0.93}\\\cline{1-4}
MMI-S2SA: \!\!&\!\!Haha, thanks. & \cellcolor[gray]{0.93} & \cellcolor[gray]{0.93} & \cellcolor[gray]{0.93} & \cellcolor[gray]{0.93} &\cellcolor[gray]{0.93}\\\cline{1-2}
DialogueGAN: \!\!&\!\!I think so. & \cellcolor[gray]{0.93} & \cellcolor[gray]{0.93} & \cellcolor[gray]{0.93} & \cellcolor[gray]{0.93} &\cellcolor[gray]{0.93}\\\cline{1-2}
DPGAN: \!\!&\!\!I'm UNK. & \cellcolor[gray]{0.93} & \cellcolor[gray]{0.93} & \cellcolor[gray]{0.93} & \cellcolor[gray]{0.93} &\cellcolor[gray]{0.93}\\\cline{1-2}\cline{5-5}
RankGAN-G: \!\!&\!\!Haha, I'm also UNK.  &\cellcolor[gray]{0.93}&\cellcolor[gray]{0.93}&$\surd$&\cellcolor[gray]{0.93} &\cellcolor[gray]{0.93}\\\cline{1-2}\cline{5-6}
ensGAN-G1: \!\!&\!\!Thank you my friend. &\cellcolor[gray]{0.93} &\cellcolor[gray]{0.93}&\cellcolor[gray]{0.93}&$\surd$  &\cellcolor[gray]{0.93}\\\cline{1-2}\cline{6-7}
Multi-Seq2Seq: \!\!&\!\! Haha. & \cellcolor[gray]{0.93} & \cellcolor[gray]{0.93} & \cellcolor[gray]{0.93} & \cellcolor[gray]{0.93} &$\surd$\\ \hline\hline
$q$: \!\!&\!\!Who's called hot pepper? & \cellcolor[gray]{0.93} & \cellcolor[gray]{0.93} & \cellcolor[gray]{0.93} & \cellcolor[gray]{0.93} &\cellcolor[gray]{0.93}\\ \cline{1-2}
$r$: \!\!&\!\!It's a girl we saw previously. & \cellcolor[gray]{0.93} & \cellcolor[gray]{0.93} & \cellcolor[gray]{0.93} & \cellcolor[gray]{0.93}&\cellcolor[gray]{0.93}\\ \hline
TF-IDF-1: \!\!&\!\!Our favourite fast food. &$\surd$ &&& &$\surd$\\ \hline
TF-IDF-2: \!\!&\!\!You don't even know hot pepper? & &$\surd$&&$\surd$  &\\ \hline
S2SA: \!\!&\!\!I know you're male.  &&& \cellcolor[gray]{0.93} & \cellcolor[gray]{0.93} &\cellcolor[gray]{0.93}\\ \cline{1-4}
MMI-S2SA: \!\!&\!\!I know you're male, I'm female. & \cellcolor[gray]{0.93} & \cellcolor[gray]{0.93} & \cellcolor[gray]{0.93} & \cellcolor[gray]{0.93} &\cellcolor[gray]{0.93}\\\cline{1-2}
DialogueGAN: \!\!&\!\!Let me guess who you are. & \cellcolor[gray]{0.93} & \cellcolor[gray]{0.93} & \cellcolor[gray]{0.93} & \cellcolor[gray]{0.93} &\cellcolor[gray]{0.93}\\\cline{1-2}
DPGAN: \!\!&\!\!I mean that I'm actually her. & \cellcolor[gray]{0.93} & \cellcolor[gray]{0.93} & \cellcolor[gray]{0.93} & \cellcolor[gray]{0.93} &\cellcolor[gray]{0.93}\\\cline{1-2}\cline{5-5}
RankGAN-G: \!\!&\!\!Are you talking about your joke? &\cellcolor[gray]{0.93} &\cellcolor[gray]{0.93} &$\surd$ &\cellcolor[gray]{0.93} &\cellcolor[gray]{0.93}\\\cline{1-2}\cline{5-6}
ensGAN-G1: \!\!&\!\!I knew it was you! &\cellcolor[gray]{0.93} &\cellcolor[gray]{0.93}&\cellcolor[gray]{0.93}& &\cellcolor[gray]{0.93}\\\cline{1-2}\cline{6-7}
Multi-Seq2Seq: \!\!&\!\!Yeah, me too. & \cellcolor[gray]{0.93} & \cellcolor[gray]{0.93} & \cellcolor[gray]{0.93} & \cellcolor[gray]{0.93} &\\
\Xhline{1pt} 
\end{tabular}
}
\vspace{-4mm}
\label{tb:case}
\end{table}

\vspace{-0.5mm}
\subsubsection{Case Study}
\mytbref{tb:case} shows several example response generation by ensembles, together with various baselines. It's obvious that an ensemble becomes plausible for selecting one final response from multiple candidates in case a single approach fails to respond correctly, just as the second and third case, corresponding to generative-failure and retrieval-failure respectively. We could also observe that as for generation module, most enhanced seq2seq models are better than S2SA in terms of both language fluency and informativeness. Moreover, the GAN-enhanced seq2seq models are generally better than MMI-S2SA which generates irrelevant responses like "I know you're male" given the query "who's called hot pepper?" in the last case. Among GAN-based generators, all DPGAN, RankGAN and ensGAN achieve similar performances in terms of the generation enrichness, which seem slightly better than dialogueGAN. Besides, while the original GBDT ranker and IRGAN-D mostly prefer the retrieved candidates, RankGAN-D however largely favors synthetic responses, conforming with their respective GAN initiatives. In contrast, the ensGAN-D is able to perform more balanced and logical selections between its generation module and pre-retrieval module, demonstrating its ability to leverage both advantages of single retrieval-based and generation-based approach in dialogue generation scenarios.

\section{Conclusion and Future Work}
In this paper, we proposed a novel generative adversarial framework that aims at enhancing a conversation retrieval-generation ensemble model by unifying GAN mechanism for both generative and retrieval approaches. The ensembleGAN enables the two generators to generate responses getting higher scores from the discriminative ranker, while the discriminator scores down adversarial samples and selects responses featuring merits of both generators, allowing for both generation and retrieval-based methods to be mutually enhanced. 
Experimental results on a large STC dataset demonstrate that our ensembleGAN outperforms other GAN mechanism on both human and automatic evaluation metrics and is capable of bringing better global optimal results.


\begin{acks}
This work was supported by the National Key Research and Development Program of China (No. 2017YFC0804001), the National Science Foundation of China (NSFC Nos. 61672058 and 61876196). 
\end{acks}

\bibliographystyle{acmart}
\bibliography{acmart}


\begin{thebibliography}{41}


\ifx \showCODEN    \undefined \def \showCODEN     #1{\unskip}     \fi
\ifx \showDOI      \undefined \def \showDOI       #1{#1}\fi
\ifx \showISBNx    \undefined \def \showISBNx     #1{\unskip}     \fi
\ifx \showISBNxiii \undefined \def \showISBNxiii  #1{\unskip}     \fi
\ifx \showISSN     \undefined \def \showISSN      #1{\unskip}     \fi
\ifx \showLCCN     \undefined \def \showLCCN      #1{\unskip}     \fi
\ifx \shownote     \undefined \def \shownote      #1{#1}          \fi
\ifx \showarticletitle \undefined \def \showarticletitle #1{#1}   \fi
\ifx \showURL      \undefined \def \showURL       {\relax}        \fi
\providecommand\bibfield[2]{#2}
\providecommand\bibinfo[2]{#2}
\providecommand\natexlab[1]{#1}
\providecommand\showeprint[2][]{arXiv:#2}

\bibitem[\protect\citeauthoryear{Bahdanau, Cho, and Bengio}{Bahdanau
  et~al\mbox{.}}{2015}]%
        {DBLP:journals/corr/BahdanauCB14}
\bibfield{author}{\bibinfo{person}{Dzmitry Bahdanau},
  \bibinfo{person}{Kyunghyun Cho}, {and} \bibinfo{person}{Yoshua Bengio}.}
  \bibinfo{year}{2015}\natexlab{}.
\newblock \showarticletitle{Neural Machine Translation by Jointly Learning to
  Align and Translate}. In \bibinfo{booktitle}{\emph{ICLR}}.
\newblock


\bibitem[\protect\citeauthoryear{Chen, Liu, Yin, and Tang}{Chen
  et~al\mbox{.}}{2017}]%
        {DBLP:journals/sigkdd/ChenLYT17}
\bibfield{author}{\bibinfo{person}{Hongshen Chen}, \bibinfo{person}{Xiaorui
  Liu}, \bibinfo{person}{Dawei Yin}, {and} \bibinfo{person}{Jiliang Tang}.}
  \bibinfo{year}{2017}\natexlab{}.
\newblock \showarticletitle{A Survey on Dialogue Systems: Recent Advances and
  New Frontiers}.
\newblock \bibinfo{journal}{\emph{{SIGKDD} Explorations}} \bibinfo{volume}{19},
  \bibinfo{number}{2} (\bibinfo{year}{2017}), \bibinfo{pages}{25--35}.
\newblock


\bibitem[\protect\citeauthoryear{Dai, Fidler, Urtasun, and Lin}{Dai
  et~al\mbox{.}}{2017}]%
        {DBLP:conf/iccv/DaiFUL17}
\bibfield{author}{\bibinfo{person}{Bo Dai}, \bibinfo{person}{Sanja Fidler},
  \bibinfo{person}{Raquel Urtasun}, {and} \bibinfo{person}{Dahua Lin}.}
  \bibinfo{year}{2017}\natexlab{}.
\newblock \showarticletitle{Towards Diverse and Natural Image Descriptions via
  a Conditional {GAN}}. In \bibinfo{booktitle}{\emph{ICCV}}.
  \bibinfo{pages}{2989--2998}.
\newblock


\bibitem[\protect\citeauthoryear{Fleiss}{Fleiss}{1971}]%
        {Fleiss1971Measuring}
\bibfield{author}{\bibinfo{person}{Joseph~L Fleiss}.}
  \bibinfo{year}{1971}\natexlab{}.
\newblock \showarticletitle{Measuring nominal scale agreement among many
  raters.}
\newblock \bibinfo{journal}{\emph{Psychological Bulletin}}
  \bibinfo{volume}{76}, \bibinfo{number}{5} (\bibinfo{year}{1971}),
  \bibinfo{pages}{378--382}.
\newblock


\bibitem[\protect\citeauthoryear{Goodfellow, Pouget{-}Abadie, Mirza, Xu,
  Warde{-}Farley, Ozair, Courville, and Bengio}{Goodfellow
  et~al\mbox{.}}{2014}]%
        {DBLP:conf/nips/GoodfellowPMXWOCB14}
\bibfield{author}{\bibinfo{person}{Ian~J. Goodfellow}, \bibinfo{person}{Jean
  Pouget{-}Abadie}, \bibinfo{person}{Mehdi Mirza}, \bibinfo{person}{Bing Xu},
  \bibinfo{person}{David Warde{-}Farley}, \bibinfo{person}{Sherjil Ozair},
  \bibinfo{person}{Aaron~C. Courville}, {and} \bibinfo{person}{Yoshua Bengio}.}
  \bibinfo{year}{2014}\natexlab{}.
\newblock \showarticletitle{Generative Adversarial Nets}. In
  \bibinfo{booktitle}{\emph{NIPS}}. \bibinfo{pages}{2672--2680}.
\newblock


\bibitem[\protect\citeauthoryear{Herbrich}{Herbrich}{2008}]%
        {Herbrich2008Large}
\bibfield{author}{\bibinfo{person}{R Herbrich}.}
  \bibinfo{year}{2008}\natexlab{}.
\newblock \showarticletitle{Large margin rank boundaries for ordinal
  regression}.
\newblock \bibinfo{journal}{\emph{Advances in Large Margin Classifiers}}
  \bibinfo{volume}{88} (\bibinfo{year}{2008}).
\newblock


\bibitem[\protect\citeauthoryear{Hochreiter and Schmidhuber}{Hochreiter and
  Schmidhuber}{1997}]%
        {DBLP:journals/neco/HochreiterS97}
\bibfield{author}{\bibinfo{person}{Sepp Hochreiter} {and}
  \bibinfo{person}{J{\"{u}}rgen Schmidhuber}.} \bibinfo{year}{1997}\natexlab{}.
\newblock \showarticletitle{Long Short-Term Memory}.
\newblock \bibinfo{journal}{\emph{Neural Computation}} \bibinfo{volume}{9},
  \bibinfo{number}{8} (\bibinfo{year}{1997}), \bibinfo{pages}{1735--1780}.
\newblock


\bibitem[\protect\citeauthoryear{Kingma and Ba}{Kingma and Ba}{2015}]%
        {DBLP:journals/corr/KingmaB14}
\bibfield{author}{\bibinfo{person}{Diederik~P. Kingma} {and}
  \bibinfo{person}{Jimmy Ba}.} \bibinfo{year}{2015}\natexlab{}.
\newblock \showarticletitle{Adam: {A} Method for Stochastic Optimization}. In
  \bibinfo{booktitle}{\emph{ICLR}}.
\newblock


\bibitem[\protect\citeauthoryear{Li, Galley, Brockett, Gao, and Dolan}{Li
  et~al\mbox{.}}{2016}]%
        {DBLP:conf/naacl/LiGBGD16}
\bibfield{author}{\bibinfo{person}{Jiwei Li}, \bibinfo{person}{Michel Galley},
  \bibinfo{person}{Chris Brockett}, \bibinfo{person}{Jianfeng Gao}, {and}
  \bibinfo{person}{Bill Dolan}.} \bibinfo{year}{2016}\natexlab{}.
\newblock \showarticletitle{A Diversity-Promoting Objective Function for Neural
  Conversation Models}. In \bibinfo{booktitle}{\emph{NAACL}}.
  \bibinfo{pages}{110--119}.
\newblock


\bibitem[\protect\citeauthoryear{Li, Monroe, Shi, Jean, Ritter, and
  Jurafsky}{Li et~al\mbox{.}}{2017}]%
        {DBLP:conf/emnlp/LiMSJRJ17}
\bibfield{author}{\bibinfo{person}{Jiwei Li}, \bibinfo{person}{Will Monroe},
  \bibinfo{person}{Tianlin Shi}, \bibinfo{person}{S{\'{e}}bastien Jean},
  \bibinfo{person}{Alan Ritter}, {and} \bibinfo{person}{Dan Jurafsky}.}
  \bibinfo{year}{2017}\natexlab{}.
\newblock \showarticletitle{Adversarial Learning for Neural Dialogue
  Generation}. In \bibinfo{booktitle}{\emph{EMNLP}}.
  \bibinfo{pages}{2157--2169}.
\newblock


\bibitem[\protect\citeauthoryear{Lin, Li, He, Sun, and Zhang}{Lin
  et~al\mbox{.}}{2017}]%
        {DBLP:conf/nips/LinLHSZ17}
\bibfield{author}{\bibinfo{person}{Kevin Lin}, \bibinfo{person}{Dianqi Li},
  \bibinfo{person}{Xiaodong He}, \bibinfo{person}{Ming{-}Ting Sun}, {and}
  \bibinfo{person}{Zhengyou Zhang}.} \bibinfo{year}{2017}\natexlab{}.
\newblock \showarticletitle{Adversarial Ranking for Language Generation}. In
  \bibinfo{booktitle}{\emph{NIPS}}. \bibinfo{pages}{3158--3168}.
\newblock


\bibitem[\protect\citeauthoryear{Liu, Lowe, Serban, Noseworthy, Charlin, and
  Pineau}{Liu et~al\mbox{.}}{2016}]%
        {DBLP:conf/emnlp/LiuLSNCP16}
\bibfield{author}{\bibinfo{person}{Chia{-}Wei Liu}, \bibinfo{person}{Ryan
  Lowe}, \bibinfo{person}{Iulian Serban}, \bibinfo{person}{Michael Noseworthy},
  \bibinfo{person}{Laurent Charlin}, {and} \bibinfo{person}{Joelle Pineau}.}
  \bibinfo{year}{2016}\natexlab{}.
\newblock \showarticletitle{How {NOT} To Evaluate Your Dialogue System: An
  Empirical Study of Unsupervised Evaluation Metrics for Dialogue Response
  Generation}. In \bibinfo{booktitle}{\emph{EMNLP}}.
  \bibinfo{pages}{2122--2132}.
\newblock


\bibitem[\protect\citeauthoryear{Liu, Lu, Yang, Qu, Zhu, and Li}{Liu
  et~al\mbox{.}}{2018}]%
        {DBLP:conf/aaai/LiuLYQZL18}
\bibfield{author}{\bibinfo{person}{Linqing Liu}, \bibinfo{person}{Yao Lu},
  \bibinfo{person}{Min Yang}, \bibinfo{person}{Qiang Qu}, \bibinfo{person}{Jia
  Zhu}, {and} \bibinfo{person}{Hongyan Li}.} \bibinfo{year}{2018}\natexlab{}.
\newblock \showarticletitle{Generative Adversarial Network for Abstractive Text
  Summarization}. In \bibinfo{booktitle}{\emph{AAAI}}.
\newblock


\bibitem[\protect\citeauthoryear{Mirza and Osindero}{Mirza and
  Osindero}{2014}]%
        {DBLP:journals/corr/MirzaO14}
\bibfield{author}{\bibinfo{person}{Mehdi Mirza} {and} \bibinfo{person}{Simon
  Osindero}.} \bibinfo{year}{2014}\natexlab{}.
\newblock \showarticletitle{Conditional Generative Adversarial Nets}.
\newblock \bibinfo{journal}{\emph{CoRR}}  \bibinfo{volume}{abs/1411.1784}
  (\bibinfo{year}{2014}).
\newblock


\bibitem[\protect\citeauthoryear{Mou, Song, Yan, Li, Zhang, and Jin}{Mou
  et~al\mbox{.}}{2016}]%
        {DBLP:conf/coling/MouSYL0J16}
\bibfield{author}{\bibinfo{person}{Lili Mou}, \bibinfo{person}{Yiping Song},
  \bibinfo{person}{Rui Yan}, \bibinfo{person}{Ge Li}, \bibinfo{person}{Lu
  Zhang}, {and} \bibinfo{person}{Zhi Jin}.} \bibinfo{year}{2016}\natexlab{}.
\newblock \showarticletitle{Sequence to Backward and Forward Sequences: {A}
  Content-Introducing Approach to Generative Short-Text Conversation}. In
  \bibinfo{booktitle}{\emph{COLING}}. \bibinfo{pages}{3349--3358}.
\newblock


\bibitem[\protect\citeauthoryear{Papineni, Roukos, Ward, and Zhu}{Papineni
  et~al\mbox{.}}{2002}]%
        {papineni2002bleu}
\bibfield{author}{\bibinfo{person}{Kishore Papineni}, \bibinfo{person}{Salim
  Roukos}, \bibinfo{person}{Todd Ward}, {and} \bibinfo{person}{Wei-Jing Zhu}.}
  \bibinfo{year}{2002}\natexlab{}.
\newblock \showarticletitle{{BLEU}: a method for automatic evaluation of
  machine translation}. In \bibinfo{booktitle}{\emph{ACL}}.
  \bibinfo{pages}{311--318}.
\newblock


\bibitem[\protect\citeauthoryear{Shang, Lu, and Li}{Shang
  et~al\mbox{.}}{2015}]%
        {NRM}
\bibfield{author}{\bibinfo{person}{Lifeng Shang}, \bibinfo{person}{Zhengdong
  Lu}, {and} \bibinfo{person}{Hang Li}.} \bibinfo{year}{2015}\natexlab{}.
\newblock \showarticletitle{Neural Responding Machine for Short-Text
  Conversation}. In \bibinfo{booktitle}{\emph{ACL}}.
  \bibinfo{pages}{1577--1586}.
\newblock


\bibitem[\protect\citeauthoryear{Song, Li, Zhang, Zhao, and Yan}{Song
  et~al\mbox{.}}{2018}]%
        {DBLP:conf/ijcai/SongLNZZY18}
\bibfield{author}{\bibinfo{person}{Yiping Song}, \bibinfo{person}{Cheng{-}Te
  Li}, \bibinfo{person}{Ming Zhang}, \bibinfo{person}{Dongyan Zhao}, {and}
  \bibinfo{person}{Rui Yan}.} \bibinfo{year}{2018}\natexlab{}.
\newblock \showarticletitle{An Ensemble of Retrieval-Based and Generation-Based
  Human-Computer Conversation Systems}. In \bibinfo{booktitle}{\emph{IJCAI}}.
  \bibinfo{pages}{4382--4388}.
\newblock


\bibitem[\protect\citeauthoryear{Sutton, McAllester, Singh, and Mansour}{Sutton
  et~al\mbox{.}}{1999}]%
        {DBLP:conf/nips/SuttonMSM99}
\bibfield{author}{\bibinfo{person}{Richard~S. Sutton},
  \bibinfo{person}{David~A. McAllester}, \bibinfo{person}{Satinder~P. Singh},
  {and} \bibinfo{person}{Yishay Mansour}.} \bibinfo{year}{1999}\natexlab{}.
\newblock \showarticletitle{Policy Gradient Methods for Reinforcement Learning
  with Function Approximation}. In \bibinfo{booktitle}{\emph{NIPS}}.
  \bibinfo{pages}{1057--1063}.
\newblock


\bibitem[\protect\citeauthoryear{Tao, Gao, Shang, Wu, Zhao, and Yan}{Tao
  et~al\mbox{.}}{2018a}]%
        {tao2018get}
\bibfield{author}{\bibinfo{person}{Chongyang Tao}, \bibinfo{person}{Shen Gao},
  \bibinfo{person}{Mingyue Shang}, \bibinfo{person}{Wei Wu},
  \bibinfo{person}{Dongyan Zhao}, {and} \bibinfo{person}{Rui Yan}.}
  \bibinfo{year}{2018}\natexlab{a}.
\newblock \showarticletitle{Get the Point of My Utterance! Learning Towards
  Effective Responses with Multi-head Attention Mechanism}. In
  \bibinfo{booktitle}{\emph{IJCAI}}. \bibinfo{pages}{4418--4424}.
\newblock


\bibitem[\protect\citeauthoryear{Tao, Mou, Zhao, and Yan}{Tao
  et~al\mbox{.}}{2018b}]%
        {DBLP:conf/aaai/TaoMZY18}
\bibfield{author}{\bibinfo{person}{Chongyang Tao}, \bibinfo{person}{Lili Mou},
  \bibinfo{person}{Dongyan Zhao}, {and} \bibinfo{person}{Rui Yan}.}
  \bibinfo{year}{2018}\natexlab{b}.
\newblock \showarticletitle{{RUBER:} An Unsupervised Method for Automatic
  Evaluation of Open-Domain Dialog Systems}. In
  \bibinfo{booktitle}{\emph{AAAI}}. \bibinfo{pages}{722--729}.
\newblock


\bibitem[\protect\citeauthoryear{Tao, Wu, Xu, Hu, Zhao, and Yan}{Tao
  et~al\mbox{.}}{2019}]%
        {tao2019multi}
\bibfield{author}{\bibinfo{person}{Chongyang Tao}, \bibinfo{person}{Wei Wu},
  \bibinfo{person}{Can Xu}, \bibinfo{person}{Wenpeng Hu},
  \bibinfo{person}{Dongyan Zhao}, {and} \bibinfo{person}{Rui Yan}.}
  \bibinfo{year}{2019}\natexlab{}.
\newblock \showarticletitle{Multi-Representation Fusion Network for Multi-Turn
  Response Selection in Retrieval-Based Chatbots}. In
  \bibinfo{booktitle}{\emph{WSDM}}. \bibinfo{pages}{267--275}.
\newblock


\bibitem[\protect\citeauthoryear{Vijayakumar, Cogswell, Selvaraju, Sun, Lee,
  Crandall, and Batra}{Vijayakumar et~al\mbox{.}}{2018}]%
        {DBLP:journals/corr/VijayakumarCSSL16}
\bibfield{author}{\bibinfo{person}{Ashwin~K. Vijayakumar},
  \bibinfo{person}{Michael Cogswell}, \bibinfo{person}{Ramprasaath~R.
  Selvaraju}, \bibinfo{person}{Qing Sun}, \bibinfo{person}{Stefan Lee},
  \bibinfo{person}{David~J. Crandall}, {and} \bibinfo{person}{Dhruv Batra}.}
  \bibinfo{year}{2018}\natexlab{}.
\newblock \showarticletitle{Diverse Beam Search for Improved Description of
  Complex Scenes}. In \bibinfo{booktitle}{\emph{AAAI}}.
  \bibinfo{pages}{7371--7379}.
\newblock


\bibitem[\protect\citeauthoryear{Wang, Jiang, and Yang}{Wang
  et~al\mbox{.}}{2017a}]%
        {DBLP:conf/kdd/WangJY17}
\bibfield{author}{\bibinfo{person}{Chenglong Wang}, \bibinfo{person}{Feijun
  Jiang}, {and} \bibinfo{person}{Hongxia Yang}.}
  \bibinfo{year}{2017}\natexlab{a}.
\newblock \showarticletitle{A Hybrid Framework for Text Modeling with
  Convolutional {RNN}}. In \bibinfo{booktitle}{\emph{SIGKDD}}.
  \bibinfo{pages}{2061--2069}.
\newblock


\bibitem[\protect\citeauthoryear{Wang, Lu, Li, and Chen}{Wang
  et~al\mbox{.}}{2013}]%
        {DBLP:conf/emnlp/WangLLC13}
\bibfield{author}{\bibinfo{person}{Hao Wang}, \bibinfo{person}{Zhengdong Lu},
  \bibinfo{person}{Hang Li}, {and} \bibinfo{person}{Enhong Chen}.}
  \bibinfo{year}{2013}\natexlab{}.
\newblock \showarticletitle{A Dataset for Research on Short-Text
  Conversations}. In \bibinfo{booktitle}{\emph{EMNLP}}.
  \bibinfo{pages}{935--945}.
\newblock


\bibitem[\protect\citeauthoryear{Wang, Yu, Zhang, Gong, Xu, Wang, Zhang, and
  Zhang}{Wang et~al\mbox{.}}{2017b}]%
        {DBLP:conf/sigir/WangYZGXWZZ17}
\bibfield{author}{\bibinfo{person}{Jun Wang}, \bibinfo{person}{Lantao Yu},
  \bibinfo{person}{Weinan Zhang}, \bibinfo{person}{Yu Gong},
  \bibinfo{person}{Yinghui Xu}, \bibinfo{person}{Benyou Wang},
  \bibinfo{person}{Peng Zhang}, {and} \bibinfo{person}{Dell Zhang}.}
  \bibinfo{year}{2017}\natexlab{b}.
\newblock \showarticletitle{{IRGAN:} {A} Minimax Game for Unifying Generative
  and Discriminative Information Retrieval Models}. In
  \bibinfo{booktitle}{\emph{SIGIR}}. \bibinfo{pages}{515--524}.
\newblock


\bibitem[\protect\citeauthoryear{{Weston}, {Dinan}, and {Miller}}{{Weston}
  et~al\mbox{.}}{2018}]%
        {weston2018retrieve}
\bibfield{author}{\bibinfo{person}{J. {Weston}}, \bibinfo{person}{E. {Dinan}},
  {and} \bibinfo{person}{A.~H. {Miller}}.} \bibinfo{year}{2018}\natexlab{}.
\newblock \showarticletitle{Retrieve and Refine: Improved Sequence Generation
  Models For Dialogue}.
\newblock \bibinfo{journal}{\emph{CoRR}}  \bibinfo{volume}{abs/1808.04776}
  (\bibinfo{year}{2018}).
\newblock


\bibitem[\protect\citeauthoryear{Wu, Wei, Huang, Li, and Zhou}{Wu
  et~al\mbox{.}}{2018a}]%
        {DBLP:journals/corr/abs-1806-07042}
\bibfield{author}{\bibinfo{person}{Yu Wu}, \bibinfo{person}{Furu Wei},
  \bibinfo{person}{Shaohan Huang}, \bibinfo{person}{Zhoujun Li}, {and}
  \bibinfo{person}{Ming Zhou}.} \bibinfo{year}{2018}\natexlab{a}.
\newblock \showarticletitle{Response Generation by Context-aware Prototype
  Editing}.
\newblock \bibinfo{journal}{\emph{CoRR}}  \bibinfo{volume}{abs/1806.07042}
  (\bibinfo{year}{2018}).
\newblock


\bibitem[\protect\citeauthoryear{Wu, Wu, Xing, Zhou, and Li}{Wu
  et~al\mbox{.}}{2017}]%
        {wu2017sequential}
\bibfield{author}{\bibinfo{person}{Yu Wu}, \bibinfo{person}{Wei Wu},
  \bibinfo{person}{Chen Xing}, \bibinfo{person}{Ming Zhou}, {and}
  \bibinfo{person}{Zhoujun Li}.} \bibinfo{year}{2017}\natexlab{}.
\newblock \showarticletitle{Sequential matching network: A new architecture for
  multi-turn response selection in retrieval-based chatbots}. In
  \bibinfo{booktitle}{\emph{ACL}}. \bibinfo{pages}{496--505}.
\newblock


\bibitem[\protect\citeauthoryear{Wu, Wu, Yang, Xu, and Li}{Wu
  et~al\mbox{.}}{2018b}]%
        {DBLP:conf/aaai/WuWYXL18}
\bibfield{author}{\bibinfo{person}{Yu Wu}, \bibinfo{person}{Wei Wu},
  \bibinfo{person}{Dejian Yang}, \bibinfo{person}{Can Xu}, {and}
  \bibinfo{person}{Zhoujun Li}.} \bibinfo{year}{2018}\natexlab{b}.
\newblock \showarticletitle{Neural Response Generation With Dynamic
  Vocabularies}. In \bibinfo{booktitle}{\emph{AAAI}}.
  \bibinfo{pages}{5594--5601}.
\newblock


\bibitem[\protect\citeauthoryear{Xing, Wu, Wu, Liu, Huang, Zhou, and Ma}{Xing
  et~al\mbox{.}}{2017}]%
        {DBLP:journals/corr/XingWWLHZM16}
\bibfield{author}{\bibinfo{person}{Chen Xing}, \bibinfo{person}{Wei Wu},
  \bibinfo{person}{Yu Wu}, \bibinfo{person}{Jie Liu}, \bibinfo{person}{Yalou
  Huang}, \bibinfo{person}{Ming Zhou}, {and} \bibinfo{person}{Wei{-}Ying Ma}.}
  \bibinfo{year}{2017}\natexlab{}.
\newblock \showarticletitle{Topic Aware Neural Response Generation}. In
  \bibinfo{booktitle}{\emph{AAAI}}. \bibinfo{pages}{3351--3357}.
\newblock


\bibitem[\protect\citeauthoryear{Xu, Sun, Ren, Lin, Wei, and Li}{Xu
  et~al\mbox{.}}{2018}]%
        {DBLP:journals/corr/abs-1802-01345}
\bibfield{author}{\bibinfo{person}{Jingjing Xu}, \bibinfo{person}{Xu Sun},
  \bibinfo{person}{Xuancheng Ren}, \bibinfo{person}{Junyang Lin},
  \bibinfo{person}{Bingzhen Wei}, {and} \bibinfo{person}{Wei Li}.}
  \bibinfo{year}{2018}\natexlab{}.
\newblock \showarticletitle{{DP-GAN:} Diversity-Promoting Generative
  Adversarial Network for Generating Informative and Diversified Text}. In
  \bibinfo{booktitle}{\emph{EMNLP}}. \bibinfo{pages}{3940--3949}.
\newblock


\bibitem[\protect\citeauthoryear{Yan, Song, and Wu}{Yan et~al\mbox{.}}{2016}]%
        {YanSW16}
\bibfield{author}{\bibinfo{person}{Rui Yan}, \bibinfo{person}{Yiping Song},
  {and} \bibinfo{person}{Hua Wu}.} \bibinfo{year}{2016}\natexlab{}.
\newblock \showarticletitle{Learning to Respond with Deep Neural Networks for
  Retrieval-Based Human-Computer Conversation System}. In
  \bibinfo{booktitle}{\emph{SIGIR}}. \bibinfo{pages}{55--64}.
\newblock


\bibitem[\protect\citeauthoryear{Yan and Zhao}{Yan and Zhao}{2018}]%
        {yan2018coupled}
\bibfield{author}{\bibinfo{person}{Rui Yan} {and} \bibinfo{person}{Dongyan
  Zhao}.} \bibinfo{year}{2018}\natexlab{}.
\newblock \showarticletitle{Coupled context modeling for deep chit-chat:
  towards conversations between human and computer}. In
  \bibinfo{booktitle}{\emph{SIGKDD}}. \bibinfo{pages}{2574--2583}.
\newblock


\bibitem[\protect\citeauthoryear{Yan, Zhao, and E}{Yan et~al\mbox{.}}{2017}]%
        {yan2017joint}
\bibfield{author}{\bibinfo{person}{Rui Yan}, \bibinfo{person}{Dongyan Zhao},
  {and} \bibinfo{person}{Weinan E}.} \bibinfo{year}{2017}\natexlab{}.
\newblock \showarticletitle{Joint learning of response ranking and next
  utterance suggestion in human-computer conversation system}. In
  \bibinfo{booktitle}{\emph{SIGIR}}. \bibinfo{pages}{685--694}.
\newblock


\bibitem[\protect\citeauthoryear{Yang, Chen, Wang, and Xu}{Yang
  et~al\mbox{.}}{2018}]%
        {DBLP:conf/naacl/YangCWX18}
\bibfield{author}{\bibinfo{person}{Zhen Yang}, \bibinfo{person}{Wei Chen},
  \bibinfo{person}{Feng Wang}, {and} \bibinfo{person}{Bo Xu}.}
  \bibinfo{year}{2018}\natexlab{}.
\newblock \showarticletitle{Improving Neural Machine Translation with
  Conditional Sequence Generative Adversarial Nets}. In
  \bibinfo{booktitle}{\emph{NAACL}}. \bibinfo{pages}{1346--1355}.
\newblock


\bibitem[\protect\citeauthoryear{Yao, Zhang, Feng, Zhao, and Yan}{Yao
  et~al\mbox{.}}{2017}]%
        {yao2017towards}
\bibfield{author}{\bibinfo{person}{Lili Yao}, \bibinfo{person}{Yaoyuan Zhang},
  \bibinfo{person}{Yansong Feng}, \bibinfo{person}{Dongyan Zhao}, {and}
  \bibinfo{person}{Rui Yan}.} \bibinfo{year}{2017}\natexlab{}.
\newblock \showarticletitle{Towards Implicit Content-Introducing for Generative
  Short-Text Conversation Systems}. In \bibinfo{booktitle}{\emph{EMNLP}}.
  \bibinfo{pages}{2190--2199}.
\newblock


\bibitem[\protect\citeauthoryear{{Young}, {Cambria}, {Chaturvedi}, {Huang},
  {Zhou}, and {Biswas}}{{Young} et~al\mbox{.}}{2018}]%
        {young2018augmenting}
\bibfield{author}{\bibinfo{person}{T. {Young}}, \bibinfo{person}{E. {Cambria}},
  \bibinfo{person}{I. {Chaturvedi}}, \bibinfo{person}{M. {Huang}},
  \bibinfo{person}{H. {Zhou}}, {and} \bibinfo{person}{S. {Biswas}}.}
  \bibinfo{year}{2018}\natexlab{}.
\newblock \showarticletitle{Augmenting End-to-End Dialogue Systems with
  Commonsense Knowledge}. In \bibinfo{booktitle}{\emph{AAAI}}.
  \bibinfo{pages}{4970--4977}.
\newblock


\bibitem[\protect\citeauthoryear{Yu, Zhang, Wang, and Yu}{Yu
  et~al\mbox{.}}{2017}]%
        {DBLP:conf/aaai/YuZWY17}
\bibfield{author}{\bibinfo{person}{Lantao Yu}, \bibinfo{person}{Weinan Zhang},
  \bibinfo{person}{Jun Wang}, {and} \bibinfo{person}{Yong Yu}.}
  \bibinfo{year}{2017}\natexlab{}.
\newblock \showarticletitle{SeqGAN: Sequence Generative Adversarial Nets with
  Policy Gradient}. In \bibinfo{booktitle}{\emph{AAAI}}.
  \bibinfo{pages}{2852--2858}.
\newblock


\bibitem[\protect\citeauthoryear{Zhao, Zhao, and Eskenazi}{Zhao
  et~al\mbox{.}}{2017}]%
        {zhao2017learning}
\bibfield{author}{\bibinfo{person}{Tiancheng Zhao}, \bibinfo{person}{Ran Zhao},
  {and} \bibinfo{person}{Maxine Eskenazi}.} \bibinfo{year}{2017}\natexlab{}.
\newblock \showarticletitle{Learning Discourse-level Diversity for Neural
  Dialog Models using Conditional Variational Autoencoders}. In
  \bibinfo{booktitle}{\emph{ACL}}. \bibinfo{pages}{654--664}.
\newblock


\bibitem[\protect\citeauthoryear{Zhu, Park, Isola, and Efros}{Zhu
  et~al\mbox{.}}{2017}]%
        {DBLP:journals/corr/ZhuPIE17}
\bibfield{author}{\bibinfo{person}{Jun{-}Yan Zhu}, \bibinfo{person}{Taesung
  Park}, \bibinfo{person}{Phillip Isola}, {and} \bibinfo{person}{Alexei~A.
  Efros}.} \bibinfo{year}{2017}\natexlab{}.
\newblock \showarticletitle{Unpaired Image-to-Image Translation using
  Cycle-Consistent Adversarial Networks}. In \bibinfo{booktitle}{\emph{ICCV}},
  Vol.~\bibinfo{volume}{2223--2232}.
\newblock


\end{thebibliography}

\end{document}